\pdfoutput=1

\documentclass{article}

\usepackage{arxiv}
\usepackage[numbers]{natbib} 
\usepackage[utf8]{inputenc} 
\usepackage[T1]{fontenc}    
\usepackage{hyperref}       
\usepackage{url}            
\usepackage{booktabs}       
\usepackage{amsmath,amsfonts,bm, amssymb}
\usepackage{mathtools}      
\usepackage{nicefrac}       
\usepackage{microtype}      
\usepackage{lipsum}		
\usepackage{graphicx}
\usepackage{float}
\usepackage{subfig}
\usepackage{adjustbox}
\usepackage{doi}
\usepackage{comment}
\usepackage{multicol, multirow}
\usepackage[dvipsnames]{xcolor}

\usepackage[utf8]{inputenc} 
\usepackage[T1]{fontenc}    
\usepackage{hyperref}       
\usepackage{url}            
\usepackage{booktabs}       
\usepackage{amsfonts}       
\usepackage{nicefrac}       
\usepackage{microtype}      
\usepackage{lipsum}
\usepackage{graphicx}
\graphicspath{ {./images/} }

\title{Enhancing Uncertainty Quantification in Drug Discovery with Censored Regression Labels}

\author{
    \href{https://orcid.org/0000-0001-5598-0286}{\includegraphics[scale=0.06]{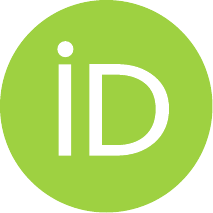}\hspace{1mm}Emma Svensson$^{1,2}$} \And \href{https://orcid.org/0000-0002-0047-0964}{\includegraphics[scale=0.06]{figures/orcid.pdf}\hspace{1mm}Hannah Rosa Friesacher$^{1,3}$} \And 
    \href{https://orcid.org/0000-0002-9808-1683}{\includegraphics[scale=0.06]{figures/orcid.pdf}\hspace{1mm}Susanne Winiwarter$^{4}$} \AND
    \href{https://orcid.org/0000-0002-7271-0824}{\includegraphics[scale=0.06]{figures/orcid.pdf}\hspace{1mm}Lewis Mervin$^{5}$} \And 
    \href{https://orcid.org/0000-0002-4901-7650}{\includegraphics[scale=0.06]{figures/orcid.pdf}\hspace{1mm}Adam Arany$^{3}$} \And 
    \href{https://orcid.org/0000-0003-4970-6461}{\includegraphics[scale=0.06]{figures/orcid.pdf}\hspace{1mm}Ola Engkvist$^{1,6}$}
    \AND
    $^1$~Molecular AI, Discovery Sciences \\ 
    AstraZeneca R\&D \\
    Gothenburg, 431 83 Sweden  \\ 
    $^5$~Cambridge, CB2 0AA UK \\
    \And
    $^2$~ELLIS Unit Linz \& \\ \textbf{Institute for Machine Learning} \\ 
    Johannes Kepler University Linz \\
    Linz, 4040 Austria 
    \And
    $^3$~ESAT-STADIUS, \\
    KU Leuven, \\
    3000 Belgium 
    \And
    $^4$~ Drug Metabolism and Pharmacokinetics, \\
    \textbf{Research and Early Development Cardiovascular,} \\
    \textbf{Renal and Metabolism (CVRM), BioPharmaceuticals R\&D,} \\ 
    AstraZeneca, Gothenburg, 431 83 Sweden \\ 
    \And
    $^6$~Department of Computer \\
    \textbf{Science and Engineering} \\ 
    Chalmers University of Technology \\ 
    Gothenburg, 412 96 Sweden
}
\date{}

\renewcommand{\headeright}{}
\renewcommand{\undertitle}{}

\hypersetup{
pdftitle={Enhancing Uncertainty Quantification in Drug Discovery with Censored Regression Labels},
pdfsubject={q-bio.NC, q-bio.QM},
pdfauthor={Emma Svensson and Hannah Rosa Friesacher and Susanne Winiwarter and Lewis Mervin and Adam Arany and Ola Engkvist},
pdfkeywords={uncertainty quantification, censored regression, temporal evaluation, distribution shift, deep learning, drug discovery, molecular property prediction},
}

\begin{document}
\maketitle              

\begin{abstract}
In the early stages of drug discovery, decisions regarding which experiments to pursue can be influenced by computational models. These decisions are critical due to the time-consuming and expensive nature of the experiments. Therefore, it is becoming essential to accurately quantify the uncertainty in machine learning predictions, such that resources can be used optimally and trust in the models improves. While computational methods for drug discovery often suffer from limited data and sparse experimental observations, additional information can exist in the form of censored labels that provide thresholds rather than precise values of observations. However, the standard approaches that quantify uncertainty in machine learning cannot fully utilize censored labels. In this work, we adapt ensemble-based, Bayesian, and Gaussian models with tools to learn from censored labels by using the Tobit model from survival analysis. Our results demonstrate that despite the partial information available in censored labels, they are essential to accurately and reliably model the real pharmaceutical setting.

\end{abstract}

\section{Introduction}
\label{sec:introduction}

    Drug discovery is a challenging field of research where experiments are time-consuming and expensive. In addition, the development of therapeutic agents bears a high risk of failure resulting in the abandonment of a drug candidate during the later stages of the drug discovery pipeline, which leads to an extensive waste of money and time. To optimize the use of resources and accelerate the drug development workflow, machine learning models are often applied to support a smart allocation of these resources \citep{mervin2021uncertainty}. In the context of machine learning-assisted drug discovery, uncertainty quantification enables safer and more reliable deployment of computational models by increasing human confidence in the models \citep{apostolakis1990concept}. The effects are highly relevant for the drug discovery pipeline as they allow users to judge results based on the predicted uncertainty quantification before deciding how to progress in the experimental workflow \citep{gal2016uncertainty}. Applying machine learning models to early-stage drug discovery requires modeling the complex chemical space where data availability is typically limited. This low-data problem is another effect of the time-consuming and costly experiments needed to generate data. As such, there is a continuously increasing need for application-specific uncertainty quantification methods in molecular property prediction, particularly in the modeling of affinity scores and drug side effects through quantitative structure-activity relationships (QSAR) \citep{hansch1964qsar}.

    Uncertainty quantification in machine learning is typically disentangled into its underlying sources, which provide a deeper understanding of the factors contributing to overall predictive uncertainty \citep{apostolakis1990concept}. The two primary sources of uncertainty are aleatoric and epistemic uncertainty \citep{kendall2017uncertainties, hullermeier2021aleatoric}. Aleatoric uncertainty refers to the inherent stochastic variability within experiments. It is often considered irreducible because it cannot be mitigated through additional data or model improvements. In drug discovery specifically, aleatoric uncertainty can reflect the inherent unpredictability of interactions between certain molecular compounds as a result of either biological stochasticity or human intervention. As such, proper quantification of aleatoric uncertainty can lead to better risk management in drug discovery by highlighting areas where outcomes are inherently uncertain \citep{yang2019analyzing}.

    On the other hand, epistemic uncertainty encompasses uncertainties related to the model’s lack of knowledge, which can stem from insufficient training data or model limitations \citep{kendall2017uncertainties, hullermeier2021aleatoric}. Unlike aleatoric uncertainty, epistemic uncertainty can be reduced by acquiring additional data or through improvements to the model. Understanding and quantifying epistemic uncertainty in drug discovery allows researchers to strategically guide data collection efforts, focusing on areas of the chemical space where the model's predictions are less certain \citep{heid2023characterizing}. By further separating these individual aspects of epistemic uncertainty, targeted enhancements to the machine learning model itself can be achieved. Recent work by \citet{gustafsson2023how} emphasizes the importance of uncertainty quantification in out-of-distribution scenarios, highlighting how epistemic uncertainty can inform about when a model will likely fail due to a lack of relevant training data. In real-world drug discovery, it is especially advantageous to understand the various sources of epistemic uncertainty, as navigating the vast chemical space efficiently and effectively can significantly impact the success of a project.
    
    Approaches that quantify uncertainty in machine learning regression tasks can be broadly categorized into several types: Bayesian learning \citep{blundell2015weight}, ensemble-based methods \citep{sheridan2012three, gal2016dropout, lakshminarayanan2017simple, scalia2020evaluating}, distance-based approaches \citep{sheridan2004similarity, berenger2018distance}, mean-variance estimation \citep{bishop1994mixture, nix1994estimating, choi2018uncertainty}, evidential learning \citep{amini2020deep}, conformal prediction \citep{vovk2005algorithmic}, and quantile regression \citep{koenker2001quantile}, among others. Recent studies have compared and benchmarked these methods on publicly available datasets for modeling of molecular properties \citep{hirschfeld2020uncertainty, kim2021bayesian, wang2021hybrid, dutschmann2023large, heid2023characterizing, yin2023evaluating}. Despite these efforts, no single method has emerged as consistently superior across all evaluation metrics and tasks \citep{yu2022uncertainty}.

    Prior research has primarily been performed on publicly available datasets using random or scaffold-based ways to split the data for evaluation. These means have been shown to either overestimate machine learning models, in the case of the random split, or underestimate the performance with the scaffold-based split \citep{hirschfeld2020uncertainty, yin2023evaluating}. \citet{hirschfeld2020uncertainty} stress the need for a more realistic evaluation, such as a temporal data split, to gain insights into the real implications and nuances between the approaches. Additionally, \citet{yin2023evaluating} point out that public benchmarks do not allow proper temporal evaluation as they lack relevant information and sufficient replications for reliable statistics. Temporal evaluation that is based on information available in public data, such as ChEMBL, for molecular property prediction can be misleading \citep{lenselink2017beyond}. The reason is that the time stamp of data points in public data relates to when the compound was added to the public domain rather than when the experiment was performed. Instead, modeling the real evolution of experiments in a pharmaceutical company is what makes a temporal evaluation truly useful. Earlier work on internal pharmaceutical assay-based data from Merck compares a temporal splitting strategy with random and structure-based splitting strategies \citep{sheridan2013time}. \citet{sheridan2013time} concludes that the temporal option best approximates the true predictive performance, but they do not explore uncertainty quantification.

    In addition to the limited amount of precise data in drug discovery, partial information in the form of censored labels is often generated during experiments. Censored labels arise when the experiment’s measurement range is exceeded, such that the exact value cannot be recorded. For instance, a fixed range of compound concentrations is typically used to test the compound's effect on biochemical processes in an assay. If no response is observed within this range, the experiment may only indicate that the response lies above or below the tested concentrations rather than providing a specific value. This results in a censored label, where the true value is known only to exceed or fall below a certain threshold. While censored labels can be easily included in classification tasks by categorizing observations as active or inactive \citep{friesacher2024towards}, integrating them into regression models that predict continuous values is far less trivial. Due to this challenge, censored data has not yet been properly utilized in regression tasks within drug discovery, despite its potential to enhance model accuracy and uncertainty quantification.

    The problem of censored regression has been widely studied in other fields, particularly in survival analysis, where \citet{hollander1987measuring} demonstrated that the information contained in data decreases after censorship. Despite the reduced information in censored labels, \citet{huttel2024bayesian} further showed that these labels are crucial in Bayesian active learning due to their contribution to increased entropy and mutual information between the predicted and true data distributions. Additionally, censored labels have enhanced quantile regression in survival analysis, as demonstrated by \citet{pearce2022censored}. In the context of drug discovery, limited work has explored ways to incorporate censored labels for regression tasks. \citet{arany2022sparsechem} proposed an adaptation of the mean squared error (MSE) to account for censored labels by using a one-sided squared loss. However, their approach did not address uncertainty quantification. To the best of our knowledge, no prior work has yet adapted both the MSE and the more general Gaussian negative log-likelihood (NLL) for uncertainty quantification in predictive modeling for drug discovery. Therefore, further exploration of how censored labels can be effectively integrated into regression models to improve prediction accuracy and uncertainty estimation in drug discovery is needed.
    
    In this work, we address the challenge of effectively utilizing censored regression labels in drug discovery, a domain where data is often scarce, and uncertainty quantification is critical for decision-making. While previous approaches have primarily focused on high-quality data, our work expands the ability of machine learning models to incorporate partial, yet valuable information provided in censored labels, thereby enhancing predictive performance and uncertainty estimation. To achieve this, we develop and extend existing methods, adapting them to the unique demands of drug discovery tasks. We also refine evaluation techniques to rigorously assess the impact of incorporating censored labels, ensuring that the benefits are clearly demonstrated across relevant metrics. Our contributions are summarized as follows:
    \begin{itemize}
        \item We derive extended versions of ensemble-based models, Bayesian models, and Gaussian mean-variance estimators capable of learning from additional partial information available in censored regression labels. 
        \item Similarly, we adapt available evaluation methods to compare models trained with and without the additional censored labels. 
        \item Furthermore, we provide a large-scale comparison between the resulting censored regression models in a comprehensive temporal evaluation using internal pharmaceutical assay-based data. 
        \item Finally, we showcase how the resulting model predictions can be used in practical applications to streamline and aid the drug discovery process in a case study.
    \end{itemize}

\section{Methods}
\label{sec:methods}
    The methodologies employed in our study are divided into two parts detailing the data and the modeling approaches. First, we describe the data used in our analysis, including the sources and types of biological assays explored. Following this, we outline the models and techniques implemented to leverage this data, focusing on how they are adapted to handle the unique challenges presented by censored regression labels.

    The analysis in this work was performed on data from 15 internal biological assays, categorized into two distinct groups: project-specific target-based assays and cross-project assays related to side effects, such as Absorption, Distribution, Metabolism, Excretion, and Toxicity (ADME-T) properties \citep{heyndrickx2023melloddy}. This division allows us to explore the effects of censored labels on uncertainty estimation using different assay types. Similar datasets were utilized in the recent study by \citet{friesacher2024towards}, providing a relevant benchmark for our work. An overview of the assays used in this study is presented in Table \ref{tab:data_overview}.

    \begin{table*}[ht]
        \caption{\textbf{Data Overview.} A detailed overview of the assays used in this analysis, including the short name used throughout the results. For each assay, key properties are shown, such as the proportion of the censored data, the end-point, the assay size, and the experimental standard deviation derived from a control compound. Due to proprietary constraints, detailed descriptions of target-based assays cannot be disclosed. The 'Goal' column in the table indicates the desired performance of a drug candidate on this assay ($\uparrow$ for high, $\downarrow$ for low values). While LogD targets can be project-specific, a target value of 2 is generally assumed, as excessively high or low values are typically undesirable.  
        }
            \label{tab:data_overview}
            \vskip 0.15in
            \centering
            \begin{adjustbox}{max width=\linewidth}
            \begin{tabular}{@{}lllcrrrc@{}}
                \toprule
                 & Assay& & & \multicolumn{1}{c}{Assay} & \multicolumn{2}{c}{Censoring} & \multicolumn{1}{c}{Control} \\
                Short Name & Description & End-point & Goal & \multicolumn{1}{c}{Size} & \multicolumn{1}{c}{$<$} & \multicolumn{1}{c}{$>$} & \multicolumn{1}{c}{Std} \\
                \midrule
                \textbf{Target-based} \\
                Target 1 & N/A & pIC50 & $\uparrow$ & 5,082 & 32\% & 1\% & 0.13 \\
                Target 2 & N/A & pIC50 & $\uparrow$ & 5,237 & 43\% & 0\% & 0.24 \\
                Target 3 & N/A & pIC50 & $\uparrow$ & 10,465 & 0\% & 0\% & 0.36 \\
                Target 4 & N/A & pIC50 & $\uparrow$ & 10,624 & 12\% & 2\% & 0.20 \\
                Target 5 & N/A & pEC50 & $\uparrow$ & 12,612 & 35\% & 0\% & 0.32 \\
                Target 6 & N/A & pIC50 & $\uparrow$ & 13,093 & 0\% & 0\% & 0.31 \\
                Target 7 & N/A & pIC50 & $\uparrow$ & 14,605 & 25\% & 0\% & 0.32 \\
                \midrule
                \textbf{ADME-T} \\
                ADME-T CYP 1 & CYP3A4 & pIC50 & $\downarrow$ & 12,875 & 61\% & 0\% & 0.11 \\ 
                ADME-T CYP 2 & CYP2C9 (a) & pIC50 & $\downarrow$ & 12,876 & 63\% & 0\% & 0.11 \\ 
                ADME-T CYP 3 & CYP2C9 (b) & pIC50 & $\downarrow$ & 14,062 & 58\% & 0\% & 0.12 \\ 
                ADME-T Perm. & Permeability & Log Perm. & $\uparrow$ & 16,511 & 8\% & 0\% & 0.17 \\  
                ADME-T Solub. & Solubility & Log Solub.  & $\uparrow$ & 47,607 & 5\% & 6\% & 0.21 \\  
                ADME-T hERG & Toxicity & pIC50 & $\downarrow$ & 67,687 & 42\% & 0\% & 0.19 \\ 
                ADME-T LogD & Lipophilicity & LogD & $\approx 2$ & 88,114 & 0\% & 8\% & 0.10 \\  
                ADME-T CLint & Metabolic Stability & Log $Cl_\text{int}$ & $\downarrow$ & 92,161 & 8\% & 6\% & 0.14 \\            
                \bottomrule
        \end{tabular}
        \end{adjustbox}
    \end{table*}

    The target-based assays are crucial for ongoing drug development projects and, therefore, cannot be disclosed fully. They all model either the half-maximal inhibitory concentration (IC50) or the half-maximal effective concentration (EC50), as seen in Table \ref{tab:data_overview}. ADME-T assays are important in the pharmaceutical industry to test the pharmacokinetic profile and safety of drug candidates. Four of the included ADME-T assays also measure IC50. Three of the included ADME-T assays measure the inhibiting effects on two isoforms of Cytochrome P450 to check for potential drug-drug interactions. Two CYP2C9 assays are included, which differ in the methods used for measuring CYP inhibition. In the CYP2C9 (b) assay, a fluorescent substrate is used to determine CYP inhibition. The two other CYP assays, CYP2C9 (a) and CYP3A4 apply an updated protocol, where the degradation of a drug is measured using liquid chromatography-mass spectrometry (LC-MS). The solubility assay determines aqueous solubility in a high throughput fashion, starting from a dimethyl sulfoxide (DMSO) stock solution where the organic solvent is evaporated to have a solid sample. The result gives the maximum concentration of a compound in an aqueous solution at pH 7.4. The included toxicity assay determines the compound’s potential to inhibit the human Ether-à-go-go-Related Gene (hERG) potassium channel. This hERG inhibition is correlated to severe cardiac side effects by prolonging the time it takes the heart to contract and relax, i.e. the QT interval. 
    
    The remaining ADME-T assays model other important properties that influence the pharmacokinetics of a drug candidate. The permeability assay measures the flux of a compound across the Caco-2 cell layer. The permeability is measured in 1E-6cm/s and is related to in vivo absorption. To obtain a compound’s lipophilicity, the logarithm of the distribution coefficient between octanol and aqueous phase at pH 7.4 is obtained (LogD). Finally, the Clint assay assesses the metabolic stability of a compound in µL/min/million cells. It measures how fast the compound is metabolized in rat hepatocytes and enables the predictions of in vivo hepatic clearance. 
    
    During the data preparation, the measurements of all target-based assays and the ADME-T assays measuring CYP and hERG inhibition were transformed to pIC50/pEC50 by taking the negative log base 10 of the measurements after transforming them to molar units. In addition, the measurements of the solubility, permeability, and Clint assays were transformed to log10-scale. The assay measuring lipophilicity was not modified, as it already comprised results on the log scale. Duplicated measurements for molecular compounds in the data were then aggregated using the median of the result. The standard deviation for this aggregation was also used in the study to indicate the experimental error. For each assay, a control compound was available which had been tested many times more than any other compound. The standard deviation of the control compound was used as an indication of the general experimental error of the assay, also presented in Table \ref{tab:data_overview}. The control compounds were removed before modeling each assay. 
    
    All remaining molecular compounds were then encoded with RDKit (version 2023.03.3)\citep{landrum2006rdkit} from SMILES strings \citep{weininger1988smiles} to Morgan Fingerprints (ECFP) \citep{morgan1965generation} of size 1024 and radius 2. Other, more advanced ways to encode molecular compounds exist, such as the graph-based ChemProp model \citep{yang2019analyzing} and the pre-trained language-based CDDD model \citep{winter2019learning}. Models based on the resulting embeddings from these neural network encoders have been compared and shown improvements in prior work \citep{lenselink2017beyond, hirschfeld2020uncertainty, dutschmann2023large}. However, \citet{dutschmann2023large} showed that fingerprints perform best in combination with random forest and are close second to CDDD in combination with neural networks. Additional prior work has similarly shown that neural networks perform better in combination with fingerprints compared to other machine learning methods \citep{mayr2018large, van2022exposing}. While we employ the fingerprint representations in this study for simplicity and computational reasons, we strongly encourage considering state-of-the-art, learned representations before deploying our proposed methods in practical applications. 

    \paragraph{Data Censoring.}
    Given that the determination of exact experimental results is often connected with additional experimental efforts, a significant proportion of the data is provided with censorship. We follow the definition of censored regression proposed by \citet{huttel2024bayesian}. Formally, if we define the true experimental label $y^*_n \in \mathbb{R}$ of a molecular compound with features $\bm{x}_n \in \mathbb{R}^d$ of length $d$ from an assay, a censored label provides only an observation of a threshold $z_n$ below or above which the true result lies. In the case of right-censoring this means that instead of observing $y^*_n$ we observe $y_n = \text{min}(y^*_n, z_n)$ and for left-censoring we instead observe $y_n = \text{max}(y^*_n, z_n)$. As such, we introduce a mask $m_n$ which is 1 if $y^*_n > z_n$ and -1 if $y^*_n < z_n$, otherwise $m_n = 0$. As a result, we get the dataset $\mathcal{D}_\text{censored} = \{\bm{x}_n, y_n, m_n\}_{n = 1}^N$ for each assay with total number of labels $N$. In particular, note that the censoring threshold can vary for different labels which is different from fixed-value censoring where the threshold is fixed \citep{powell1986censored}. Furthermore, we assume that the true distribution and the censoring distribution are conditionally independent given the molecular compound's features, i.e. that $y_n^* \perp z_n | \bm{x}_n$. Labels that are not censored are henceforth referred to as observed labels. An observed dataset was also prepared for each assay to evaluate if the performance improved by including the censored labels. These datasets contain only the observed labels defined by $\mathcal{D}_\text{observed} = \{\bm{x}_n, y_n, m_n\}_{\forall n : m_n = 0}$. 
    
    While prior work mentions the possibility of extending the definition of censored regression to include both right- and left-censoring at once \citep{arany2022sparsechem, huttel2024bayesian}, the majority of previous applications focus only on right-censoring. During the aggregation of duplicated measurements in the data preparation for this work, observed labels were always prioritized over censored labels. If no observed labels were available, the compounds were assigned the most common censored threshold among the potential duplicated measurements. Table \ref{tab:data_overview} lists the percentage of censored labels for each assay used in this work divided between left- and right-censoring, according to the log-scale of the end-points. Note that while many assays include only left-censored labels, i.e., $y^*_n < z_n$, two assays have a more balanced amount of censored labels, namely the ADME-T assays for solubility and metabolic stability. There is also one of the included assays that only have right-censored labels, namely the ADME-T assay for lipophilicity. Finally, two of the target-based assays, Target 3 and Target 6, do not have any censored labels. These two assays are useful in the model comparison to determine how all of the considered models compare when no censored labels are available. 

\begin{figure}[t]
        \centering
        \includegraphics[width=0.7\linewidth]{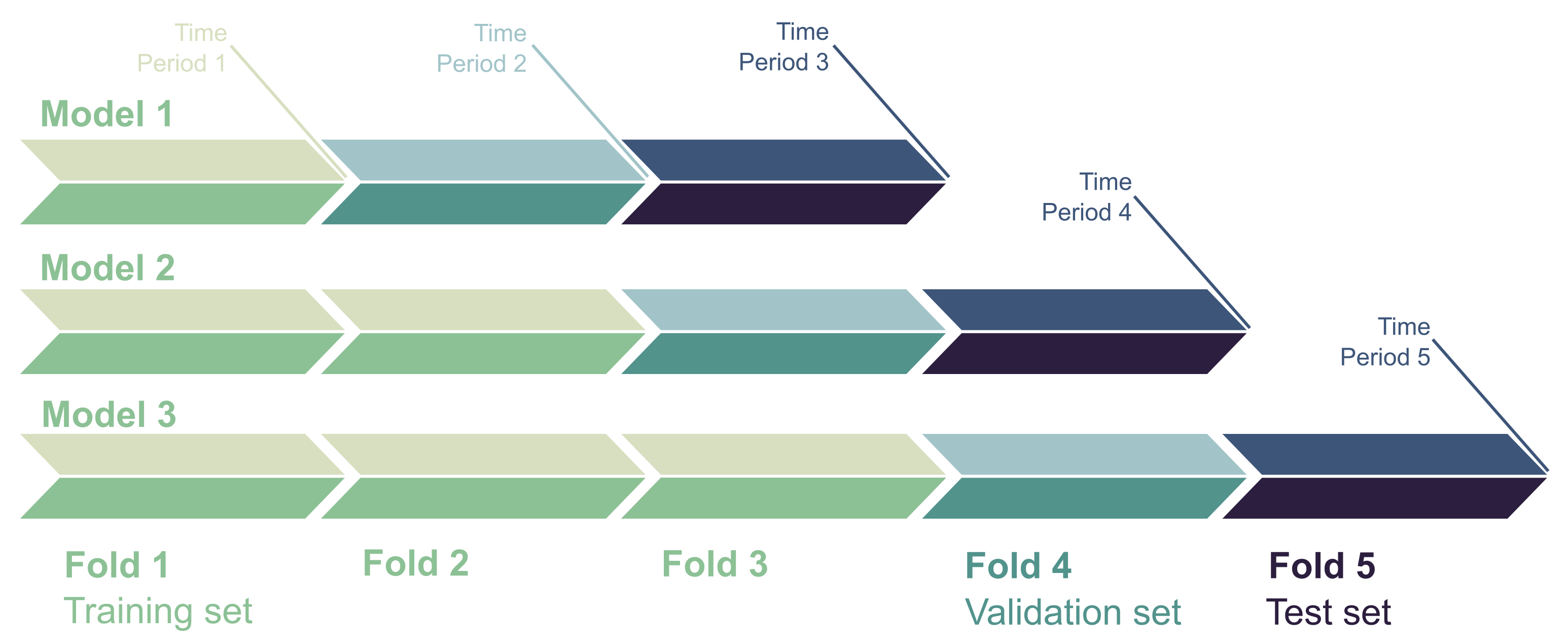}
        \caption{\textbf{Five-fold Temporal Split.} Illustrating the five folds created based on the date of the experiments and how they are used to create three temporal settings, each with more training data. For each setting, the first subsequent fold is used for validation, and the second subsequent fold is used for testing. Note that the folds were created to have roughly equal size, not based on fixed time intervals. 
        }
        \label{fig:temporal_split}
    \end{figure}

    \paragraph{Temporal Split.}
    Apart from the modeling of censored data, a key contribution of our work relates to evaluating the uncertainty quantification of molecular property prediction in a temporal setting. As such, we simulated realistic assay-based modeling of pharmaceutical projects by splitting the data of each assay into five folds based on the date of the experiment. Where duplicated measurements were aggregated, the first experiment date of all measurements was used. Fig.~\ref{fig:temporal_split} illustrates the folds and resulting three settings used to evaluate and compare trained models as time evolves in this work. The time intervals were chosen to create roughly equally sized folds regarding the number of observed labels. 

    By comparing the feature-space and label distributions between the three training folds and their respective validation and test sets, we identified various scenarios of in- versus out-of-distribution cases among the datasets. For the feature-space analysis, we created t-SNE projections of the molecular compounds in each dataset shown in Fig.~\ref{fig:tsne_distributions} in Appendix \ref{app:distributions}. For the label-space analysis, we compared the distributions of all observed labels of each fold as shown in Fig.~\ref{fig:label_distributions} in Appendix \ref{app:distributions}.
    
    Generally, for the ADME-T assays no clear distribution shift is present in either the feature-space or the label-space over time. On the contrary, for the target-based assays both the feature-space and the label-space shift over time. These observed trends are naturally expected due to the fundamental difference between the two kinds of assays. ADME-T assays include final drug candidates from all kinds of drug discovery projects that stem from a diverse chemical space. Target-based assays, on the other hand, are more focused on a given chemical space, which can shift over time as the focus of the project changes. This important distinction would not be possible without the temporal split proposed in this work, and the effects of it on model performance, as well as calibration of predicted uncertainty, is explored during the evaluation of the models. 

\subsection{Models for Uncertainty Quantification}

\begin{figure}[t]
        \centering
        \includegraphics[width=0.75\textwidth]{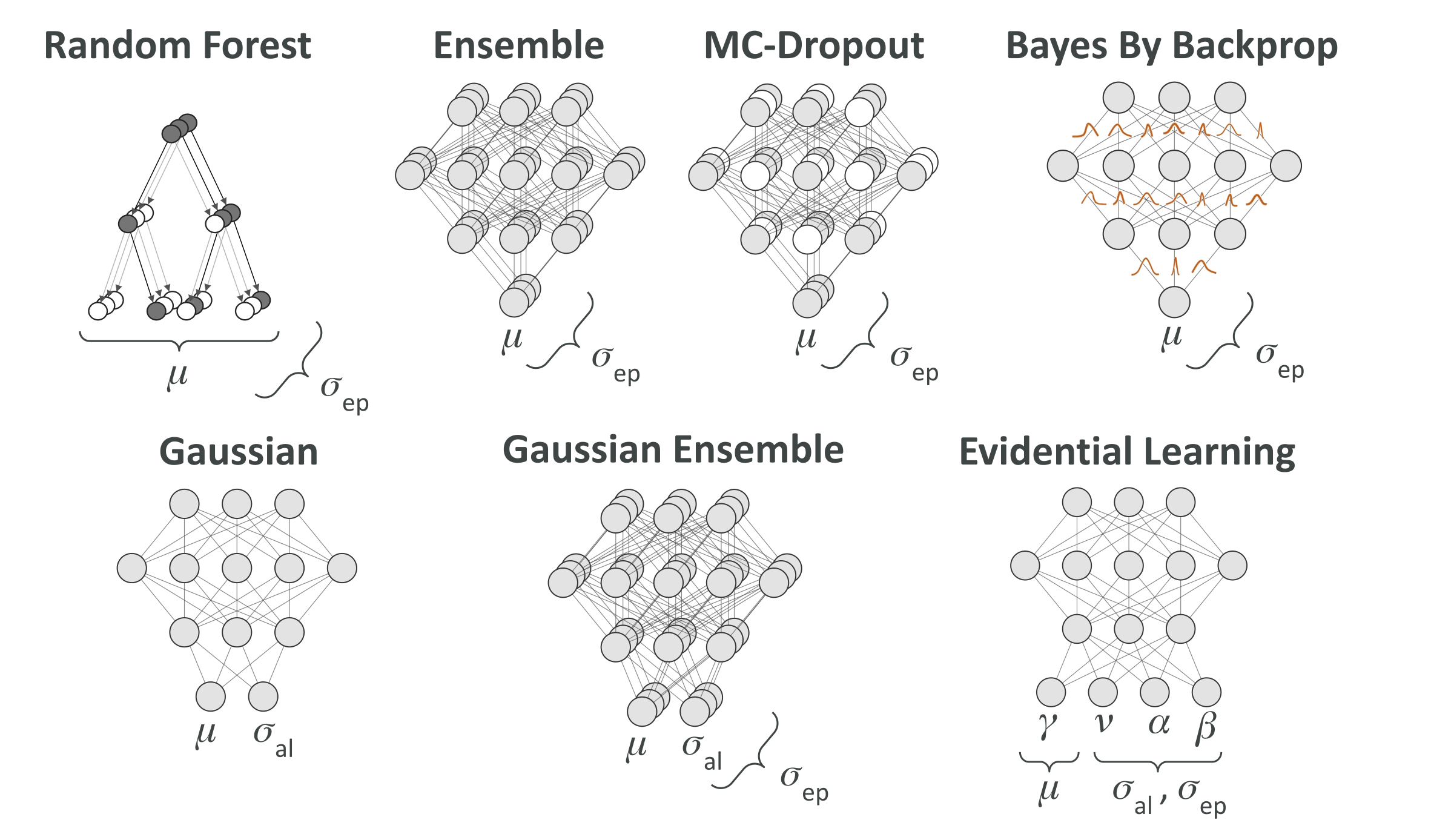}
        \caption{\textbf{Overview of Models.} Illustrations of all models used in this study. The top row shows ensemble-based and Bayesian methods for which epistemic uncertainty can be obtained from the standard deviation in sampled predictions. All models in the bottom row produce aleatoric estimates of uncertainty. Additionally, epistemic uncertainty can be derived from the Gaussian Ensemble and the Evidential model.}
        \label{fig:methods}
    \end{figure}

    In this work, we propose five models that extend existing techniques to handle censored labels alongside two baseline models that only learn from observed labels without accounting for censorship. Fig.~\ref{fig:methods} gives an overview of all models and how they can estimate epistemic and/or aleatoric uncertainty. Ensemble-based and Bayesian approaches, such as Random Forests \citep{sheridan2012three}, ensembles of neural networks \citep{lakshminarayanan2017simple}, Monte Carlo (MC) Dropout \citep{gal2016dropout} and the approximation of a Bayesian neural network known as Bayes by Backprop \citep{blundell2015weight} are commonly used to estimate epistemic uncertainty. For aleatoric uncertainty, two primary methods are available: the Gaussian mean-variance estimator \citep{nix1994estimating} and evidential deep learning \citep{amini2020deep}. By creating an ensemble of the Gaussian model, we can also derive estimates of epistemic uncertainty. Evidential deep learning, on the other hand, simultaneously models four parameters that can be used to directly estimate aleatoric and epistemic uncertainty. We extend all of these models, except the Random Forest and Evidential model, to learn from both observed and censored data, while the two remaining approaches are used as baselines. A detailed description of how the model selection is performed for each model is provided in Appendix \ref{app:model_selection}. While the data used in this work is proprietary and cannot be disclosed, the full methodology will be made available on GitHub.

    \paragraph{Ensemble-based approaches.}
    In general, ensemble-based approaches are useful to model epistemic uncertainty by estimating the model variance. An ensemble is defined as a set of $K$ base estimators $f(\bm{x}_n)$, for which the average of the individual base estimators' predictions is taken as the final prediction by the ensemble and the variance of the predictions as an estimate of the predictive uncertainty as follows,
    \begin{equation}
        \mu_n = \frac{1}{K}\sum_{k=1}^K f_k(\bm{x}_n), \quad
        \sigma^2_{\text{ep}, n} = \frac{1}{K}\sum_{k=1}^K (f_k(\bm{x}_n) - \mu_n)^2.
    \end{equation}
    
    In a Bayesian framework, the uncertainty in model parameters $\bm{\theta}$ results in the predictive uncertainty of the model $p(y_n | \bm{x}_n, \bm{\theta})$. The true posterior distribution of the model parameters for a given dataset can be described as $p(\bm{\theta} | \mathcal{D})$, such that the predictive uncertainty of the Bayesian model average is defined by $p(y_n | \bm{x}_n, \mathcal{D}) = \int_{\bm{\Theta}} p(y_n | \bm{x}_n, \Tilde{\bm{\theta}}) p(\Tilde{\bm{\theta}}|\mathcal{D}) \text{d}\Tilde{\bm{\theta}}$ \citep{gal2016uncertainty,hullermeier2021aleatoric}. As shown by \citet{lakshminarayanan2017simple} and \citet{gal2016uncertainty}, the variance of ensemble predictions approximates the epistemic part of this true posterior distribution. 

    One of the simplest and most widely used traditional machine learning methods for ensemble-based uncertainty quantification takes the decision trees in a Random Forest as an ensemble \citep{sheridan2012three}. We use the implementation of decision tree regressors from Scikit-learn \citep{scikit-learn} where each decision tree is trained independently using different sub-samples of the training data. The final prediction is obtained by averaging the predictions of all the trees and the variance constitutes the epistemic uncertainty estimate. Similarly, ensembles can be created by training multiple instances of the same neural network architecture with different random weight initialization \citep{lakshminarayanan2017simple}. We first consider the neural network architecture with a single output, which means that each base estimator makes a prediction such that the final prediction by the ensemble again is the average of the individual predictions and the variance estimates the epistemic uncertainty \citep{tetko2008critical}. We use 50 randomly initialized and independently trained neural networks to create this ensemble and hence refer to it as Ensemble. All neural networks in this work are implemented and trained with PyTorch \citep{paszke2019pytorch}.

    Furthermore, MC-Dropout is another technique that uses dropout in neural networks to approximate Bayesian inference \citep{gal2016dropout}. Dropout is typically used when training neural networks to prevent overfitting by randomly setting a fraction of the neurons to zero. During MC-Dropout, multiple samples are drawn from the model using dropout during the inference phase. Similar to the ensemble-based approaches, the variance between these predictions provides an estimate of the epistemic uncertainty. We use the same neural network architecture from the Ensemble in our MC-Dropout model but train only one model and draw 500 samples from it during the inference step. 
    
    In a standard regression framework, the mentioned base estimators of these models are trained using the MSE loss. We propose to use the adapted CensoredMSE loss from \citet{arany2022sparsechem} to allow the models to learn also from additional censored labels where available. As such, we define a one-sided error of a given prediction $\mu_n$ and the true label $y_n$ as follows,
    \begin{equation}
    \label{eq:censored_error}
        \varepsilon_n = 
        \begin{cases} 
        \text{min}\left(z_n - \mu_n, 0\right), &\text{ if } ~m_n = -1, \\
        y_n - \mu_n, &\text{ if } ~m_n = 0, \\
        \text{max}\left(z_n - \mu_n, 0\right), &\text{ if } ~m_n = 1,
        \end{cases}
    \end{equation}
    and get the CensoredMSE as $\mathcal{L}^{\text{MSE}} = \frac{1}{N} \sum_{n=1}^N \varepsilon_n^2$. The interpretation of the proposed loss is that when the model correctly predicts a value above the threshold of a right-censored label or below the threshold of a left-censored label the error is set to zero. This reflects a perfect prediction and does not result in any weight updates due to the zero gradient of a constant.

    \paragraph{Bayes by Backprop.}
    \citet{blundell2015weight} proposed a Bayesian version of neural networks, known as Bayes by Backprop, that approximates the posterior distribution of the network weights. In traditional neural networks, the weights are deterministic, meaning that a single set of values is learned during training. However, in Bayes by Backprop, the weights are modeled as random variables with associated probability distributions, typically Gaussian. This allows the network to maintain a distribution over the weights, reflecting the uncertainty about the true values of these parameters given the observed data.

    To train the Bayes by Backprop model, the goal is to approximate the posterior distribution $p(\bm{w} | \mathcal{D})$ of the weights $\bm{w}$ given the training data $\mathcal{D}$. Since exact Bayesian inference is intractable for neural networks, a variational approximation is used. The approach involves defining a variational distribution $q(\bm{w} | \bm{\theta})$, parameterized by $\bm{\theta}$, which approximates the true posterior. The optimal variational parameters $\bm{\theta}$ are learned by minimizing the variational free energy, also known as the evidence lower bound (ELBO), which balances the trade-off between fitting the data well and staying close to the prior distribution $p(\bm{w})$. The loss function used to train the Bayes by Backprop model is therefore,
    \begin{equation}
        \mathcal{L}^\text{BNN} = \text{KL} [q(\bm{w}|\bm{\theta}) || p(\bm{w})] - \mathbb{E}_{q(\bm{w}|\bm{\theta})} [\text{log} ~p(\mathcal{D}|\bm{w})],
    \end{equation}
    where the first term is the Kullback-Leibler (KL) divergence between the variational distribution $q(\bm{w} | \bm{\theta})$ and the prior distribution $p(\bm{w})$ and the second term is the expected log-likelihood of the data under the variational distribution. The KL divergence penalizes deviations from the prior distribution while the expected log-likelihood encourages the model to find weight distributions that explain the observed data well.
    
    In our implementation, we convert the same neural network used in the ensemble-based approaches into a Bayesian neural network and use the CensoredMSE as the log-likelihood to allow the Bayes by Backprop model to learn from censored labels. The KL divergence is unaffected by censored labels as it relates solely to the learned weight distribution with respect to the prior. During inference, we sample 500 sets of weights from the learned distribution, and the resulting ensemble of models generates a distribution of predictions. As for the ensemble-based approaches, the spread of these predictions provides an estimate of the epistemic uncertainty, which reflects the model's uncertainty due to limited data or model capacity.

    \paragraph{Gaussian Models.}
    So far, the base estimators of all described models produce only a prediction of the regression label. Next, we introduce the mean-variance-estimator originally proposed by \citet{nix1994estimating}. This is a neural network with two outputs: one representing the predicted value $\mu_n$ and the other representing the model's estimate of the aleatoric uncertainty $\sigma^2_{\text{al}, n}$. The predicted variance is processed through a Softplus function to ensure non-negativity and stability. This model is trained using the NLL, which in the case of a presumed Gaussian distribution becomes the following,
    \begin{equation}
        \mathcal{L}^\text{NLL} = \frac{1}{N} \sum_{n=1}^N -\text{log}\varphi(y_n | \bm{x}_n, \bm{\theta})  = \frac{1}{N} \sum_{n=1}^N
        \frac{1}{2} \text{log}(\sigma^2_{\text{al}, n}) + \frac{(y_n - \mu_n)^2}{2 \sigma^2_{\text{al}, n}} + \text{constant},
    \end{equation}
    where $\varphi(y_n | \bm{x}_n, \bm{\theta})$ is the probability density function for the Gaussian distribution with parameters $\mu_n$ and $\sigma^2_{\text{al}, n}$ estimated by the model with parameters $\bm{\theta}$. Note that the constant term $\frac{1}{2} \text{log}(2\pi)$ is omitted from the loss. Additionally, note that the variance is lower bound by a small number ($\xi = 1e^{-6}$) for numerical stability as done in the PyTorch implementation of the Gaussian NLL \citep{paszke2019pytorch}. We refer to this model as Gaussian going forward and additionally make a Gaussian Ensemble from 5 independently trained Gaussian models. For the Gaussian Ensemble we get the aleatoric estimate from the expected value of all predicted variances and the epistemic estimate from the variance of the prediction, as for the previous ensemble-based approaches. 

    To adapt this learning objective to be able to handle additional censored labels where available, we take inspiration from the Tobit model \citep{tobin1958estimation} used more recently in Bayesian active learning for survival analysis by \citet{huttel2024bayesian}. The intuition behind this method is that for censored labels we use the integral of the probability density function that is within the correct censored region. Thus, for a left-censored label with threshold $z_n$ the objective is to maximize the integral of $\varphi(y_n | \bm{x}_n, \bm{\theta})$ below $z_n$, in other words to minimize the negative logarithm of the cumulative distribution function $\Phi(z_n | \bm{x}_n, \bm{\theta})$. On the contrary, for right-censored labels with threshold $z_n$ the objective is to minimize the negative logarithm of the complement to the cumulative distribution function $1 - \Phi(z_n | \bm{x}_n, \bm{\theta})$. In summary, our proposed CensoredNLL is defined as,

    \begin{equation}
    \label{eq:nll}
        \mathcal{L}^\text{NLL} = - \frac{1}{N} \sum_{n=1}^N (1-|m_n|) ~\text{log}\varphi(y_n | \bm{x}_n, \bm{\theta}) + |m_n| ~\text{log}
        \begin{cases}
            \Phi(z_n | \bm{x}_n, \bm{\theta}), &\text{ if } ~m_n = -1, \\
            1 - \Phi(z_n | \bm{x}_n, \bm{\theta}), &\text{ if } ~m_n = 1.
        \end{cases}
    \end{equation}

    \paragraph{Evidential Deep Learning.}
    Another probabilistic approach to uncertainty estimation is through evidential learning using a neural network as proposed by \citep{amini2020deep}. This approach leverages the principles of evidential reasoning, where the network outputs parameters of a distribution that characterizes both aleatoric and epistemic uncertainty. In evidential deep learning, the neural network outputs four parameters $\gamma_n \in \mathbb{R}, \nu_n > 0, \alpha_n > 1, \beta_n > 0$, which together model the prediction $\mu_n \sim \mathcal{N}(\gamma_n, \sigma^2_n \nu_n^{-1})$ and variance $\sigma^2_n \sim \Gamma^{-1}(\alpha_n, \beta_n)$. In the implementation, the outputs for $\nu_n, \alpha_n$, and $\beta_n$ are all passed through a Softplus function, and then 1 is added to $\alpha_n$. Through this hierarchical model, the parameters express the epistemic uncertainty and aleatoric uncertainty estimates as,
    \begin{equation}
        \sigma^2_{\text{ep}, n} = \frac{\beta_n}{\nu_n(\alpha_n -1)}, \quad \sigma^2_{\text{al}, n} = \frac{\beta_n}{\alpha_n -1}.
    \end{equation}

    The Evidential model is trained using the evidential loss, which optimizes the predicted parameter $\gamma_n, \nu_n, \alpha_n, \beta_n$ to balance their fit to the data and simultaneous calibration of uncertainty estimates. The loss is defined as,
    \begin{align}
        \label{eq:el_1}
        \mathcal{L}^\text{EL} = \frac{1}{N} \sum_{n=1}^N &\frac{1}{2}~\text{log}\left(\frac{\pi}{\nu_n}\right) 
        + \text{log}\left(\frac{\Gamma(\alpha_n)}{\Gamma(\alpha_n + \frac{1}{2})}\right) - \alpha_n ~\text{log} \left(2\beta_n(1+\nu_n)\right) \\ 
        \label{eq:el_2}
        &+ \left(\alpha_n + \frac{1}{2}\right) \text{log}\left((y_n - \gamma_n)^2 \nu_n + 2\beta_n (1+\nu_n)\right) \\ 
        \label{eq:el_3}
        &+ \lambda |y_n - \gamma_n| \cdot (2\nu_n + \alpha_n), 
    \end{align}
    where $\Gamma$ is the Gamma distribution and $\lambda$ is a hyperparameter determining the level of regularization. The first part of the evidential loss (eq. \ref{eq:el_1}) corresponds to the log-likelihood of the predicted mean and variance of the normal-inverse Gamma distribution. The second part (eq. \ref{eq:el_2}) optimizes the predictions toward the true target values $y_n$ while also accounting for the uncertainty in the predictions. Finally, the last part (eq. \ref{eq:el_3}) provides the regularization that penalizes the model for being overly confident, we use $\lambda = 1$. 

    While evidential learning has been used previously to estimate uncertainty in drug discovery-related tasks \citep{wang2023uncertainty, yin2023evaluating} its effectiveness at disentangling aleatoric and epistemic uncertainty has recently been questioned \citet{juergens2024is}, therefore we decided not to provide an extension of this model for cases with censored labels. Instead, the Evidential model is used solely as a baseline to benchmark the performances of the other approaches with that of this commonly used model.

    \subsection{Evaluation.}
        The focus of this work is to determine the performance of models on realistic pharmaceutical data. As such all models are evaluated on the full test sets containing observed and, if available, censored labels. The MSE loss is used to evaluate the performance of the predictions made by the models, in cases of censored labels we use the one-sided squared error from eq. \ref{eq:censored_error}. Other metrics are required to evaluate the accuracy and calibration of the predicted uncertainties. We consider two ways to evaluate predicted uncertainty: one that evaluates only the accuracy or calibration of the uncertainty and another that evaluates predictive performance intertwined with how well-calibrated the predicted uncertainty is. 
        
        A detailed way to evaluate the predicted uncertainties by themselves is by comparing the confidence-based calibration curve to the identity function, which corresponds to perfect calibration \citep{hubschneider2019calibrating, wang2021hybrid, heid2023characterizing, yang2023explainable}. The confidence-based calibration curve is obtained by computing the z\% confidence interval (CI) for every predicted uncertainty in the test set. Next, the observed fraction of errors within each CI is calculated for several expected fractions between 0 and 1. Again, we adopt this evaluation technique for censored labels by using the one-sided squared error from eq. \ref{eq:censored_error}. 
            
        Furthermore, the Gaussian NLL \citep{zadrozny2001obtaining} and the Expected Normalized Calibration Error (ENCE) \citep{levi2022evaluating} are two global metrics that evaluate the intertwined predictive performance and calibration of uncertainties. We use the extended version of the Gaussian NLL from the Tobit model that can handle censored labels as defined in eq. \ref{eq:nll}, and does not omit any constant terms. The ENCE metric is derived from the error-based calibration plot proposed by \citet{levi2022evaluating}, which is made from a binned representation of the root MSE (RMSE) and the root mean variance (RMV), i.e., predicted uncertainty. In the case of censored labels, we again employ the one-sided squared error before taking the root of the mean. Computationally, the errors and corresponding predicted uncertainties are ordered based on increasing predicted uncertainty and split into a set $\mathcal{B}$ of bins. For each bin $b$ of size $|b|$ the RMSE and RMV are calculated as, 
        \begin{equation}
            \text{RMSE}_b = \sqrt{\frac{1}{|b|}\sum_{i \in b} (y_i - \mu_i))^2}, \quad 
            \text{RMV}_b = \sqrt{\frac{1}{|b|}\sum_{i \in b} \sigma^2_i}.
        \end{equation}
        The bins are then summarized to give the ENCE metric as follows, 
        \begin{equation}
            \text{ENCE} = \frac{1}{|\mathcal{B}|}\sum_{b \in \mathcal{B}} \frac{|\text{RMSE}_b - \text{RMV}_b|}{\text{RMV}_b}.
        \end{equation}
            
        Other metrics have been proposed and used to evaluate uncertainty estimates in drug discovery applications, such as Spearman's Rank Correlation Coefficient between predicted uncertainties and corresponding errors \citep{hirschfeld2020uncertainty, wang2021hybrid, dutschmann2023large, yin2023evaluating}. However, this score has been criticized due to the stochasticity and unreliability of the result \citet{rasmussen2023uncertain}. Statistically, a data point with high predicted uncertainty can still result in a prediction with low error and vice versa. Additionally, it is non-trivial how to adjust this score for censored labels. Therefore, we discard the metric from our analysis.

\section{Experiments}
\label{sec:experiments}

    We start our experiments by evaluating the effect of training models with the additional partial information available in the real pharmaceutical assay-based setup in the form of censored regression labels. This comparison is conducted as an ablation study, comparing the difference in overall performance between models trained with and without censored data for each assay. Next, we provide an in-depth comparison between the models on each temporal setting of every assay. The model comparison evaluates the predictive performance and calibration separately for aleatoric and epistemic uncertainty estimates. Finally, we provide practical illustrations of the resulting uncertainty estimates for the best-performing model in terms of the aleatoric and epistemic parts respectively. 

    \subsection{Ablation Study.}
        To determine the impact of training uncertainty-aware machine learning models with additional partial information in the form of censored regression labels for molecular property prediction, we compare the NLL of each model trained with and without the censored data. Each test set is kept the same, containing all available data including the censored labels. However, the baseline models are trained only on the observed labels, i.e. $\mathcal{D}_\text{observed}$. This also includes the validation set used for hyper-parameter optimization and early stopping. We define the test score from the model trained without censored data as $\text{NLL}_\text{observed}$ and the test score from the model trained with censored data as $\text{NLL}_\text{censored}$. As NLL is minimized for more accurate predictions and better-calibrated uncertainty estimates, we define the evaluation metric as $\Delta \text{NLL} = \text{NLL}_\text{observed} - \text{NLL}_\text{censored}$ such that a positive difference indicates that the model trained with censored data performed better. 
        
\begin{figure}[t]
            \centering
            \includegraphics[width=\textwidth]{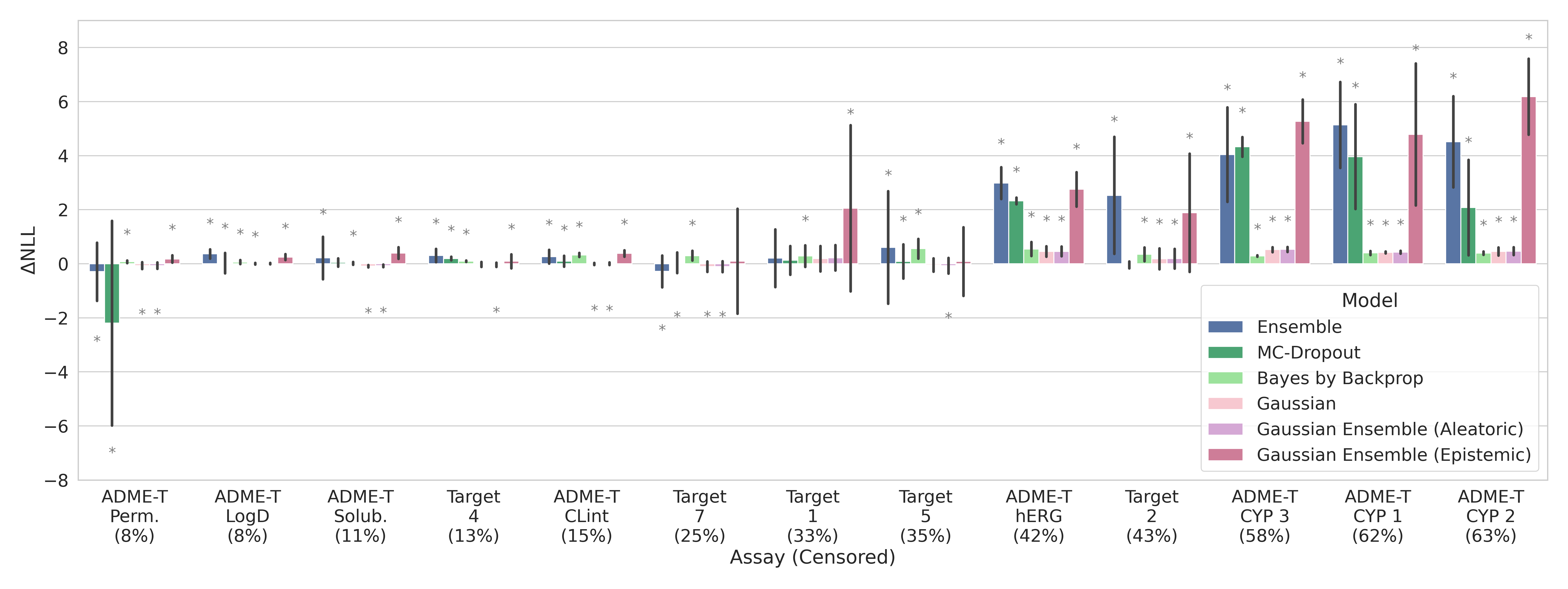}
            \caption{\textbf{Ablation Study.} Summary of the difference between NLL for models trained without censored labels and with censored labels. All three temporal settings are combined for readability. Stars above the bars indicate that the censored model was significantly better ($p<0.05$) for a majority of temporal settings, whereas stars below the bars indicate that the model trained without censored labels was significantly better for a majority of temporal settings.}
            \label{fig:ablation}
        \end{figure}

        We evaluate this difference for each assay and temporal setting and test the statistical significance using a Mann-Whitney-Wilcoxon test. If the difference was positive and significant ($p<0.05$) the model trained with censored data was best. Similarly, if the difference was negative and significant ($p<0.05$) the model trained without censored data was the best. The results of the ablation study are summarized in Fig.~\ref{fig:ablation}, and more detailed results are available in Fig.~\ref{fig:full_ablation_1}, \ref{fig:full_ablation_2}, and \ref{fig:full_ablation_3} in Appendix \ref{app:ablation}. The assays have been ordered according to the overall percentage of censored labels, with the least amount of available censored labels to the left. Note that the two assays without any available censored labels, Target 3 and Target 6, have been excluded for natural reasons. Stars indicate if the model trained with or without censored data was significantly better in a majority of the three temporal settings, and the error bars indicate the standard deviation between the temporal settings.  

        The result in Fig.~\ref{fig:ablation} shows that the performance is overwhelmingly favorable for the models trained with the censored regression labels. Especially for the models that estimate epistemic uncertainty, there are only two assays where the Ensemble and MC-Dropout models performed significantly better without censored labels. Among all other models and assays the performance was significantly better or comparable when utilizing the censored labels. In the case of aleatoric uncertainty estimates, the inclusion of censored labels during training predominately enhances the models trained on datasets with a large enough proportion of censored labels compared to observed labels, e.g. above 35\%. Still, for some assays with a smaller percentage of censored labels available, the performance is comparable to or improved by training with the censored labels. 
        
        Note that the Gaussian Ensemble is present twice in Fig.~\ref{fig:ablation} as it produces both, an estimate of the aleatoric and epistemic parts of the uncertainty respectively. As such, we conclude that the trend is not likely related to the loss functions used but rather a difference between the types of uncertainties. Recall that the Ensemble, MC-Dropout, and Bayes by Backprop models are adapted using the one-sided squared loss whereas the two Gaussian models use the Tobit loss. It is possible that the Tobit model cannot reflect the aleatoric uncertainty as accurately and that a weighting scheme would be needed in the loss to make the censored model favorable also for datasets with only a lower percentage of censored labels available. However, given the overall majority of improvements seen with censored labels included during training, we recommend the community utilize this data in the future when modeling molecular property prediction with regression. Furthermore, we continue to compare these models to each other and the additional baselines in the following section.  

    \subsection{Model Comparison.}
        We compare the proposed models trained with all available data, including censored labels, on each dataset by obtaining their predicted accuracy in terms of MSE, their local calibration in terms of the confidence-based calibration curves, and their overall performance in terms of NLL and ENCE. Where uncertainty estimates are evaluated, we separate the aleatoric and epistemic parts. For each dataset, the three temporal settings are evaluated separately such that the results can be compared between different sizes of training sets, and as time evolves. For instance, it is relevant to know if conclusions drawn from the first temporal setting also hold throughout time, i.e. on the following temporal settings. In this analysis, the assays are first ordered by category and then according to the overall size of the dataset. For the scores, MSE, NLL, and ENCE the model with the best average score over 10 repeated experiments is marked with a star. Additionally, any other performances not significantly different from the best model according to a Mann-Whitney-Wilcoxon test are also marked with stars. The two target-based assays without censored labels are included. Apart from the proposed extended models that are trained with censored labels two additional baselines are included, a Random Forest model and a model trained with evidential deep learning. These models are trained without censored labels as explained in Section \ref{sec:methods}. 

        \begin{figure}[t]
            \centering
            \includegraphics[width=\textwidth]{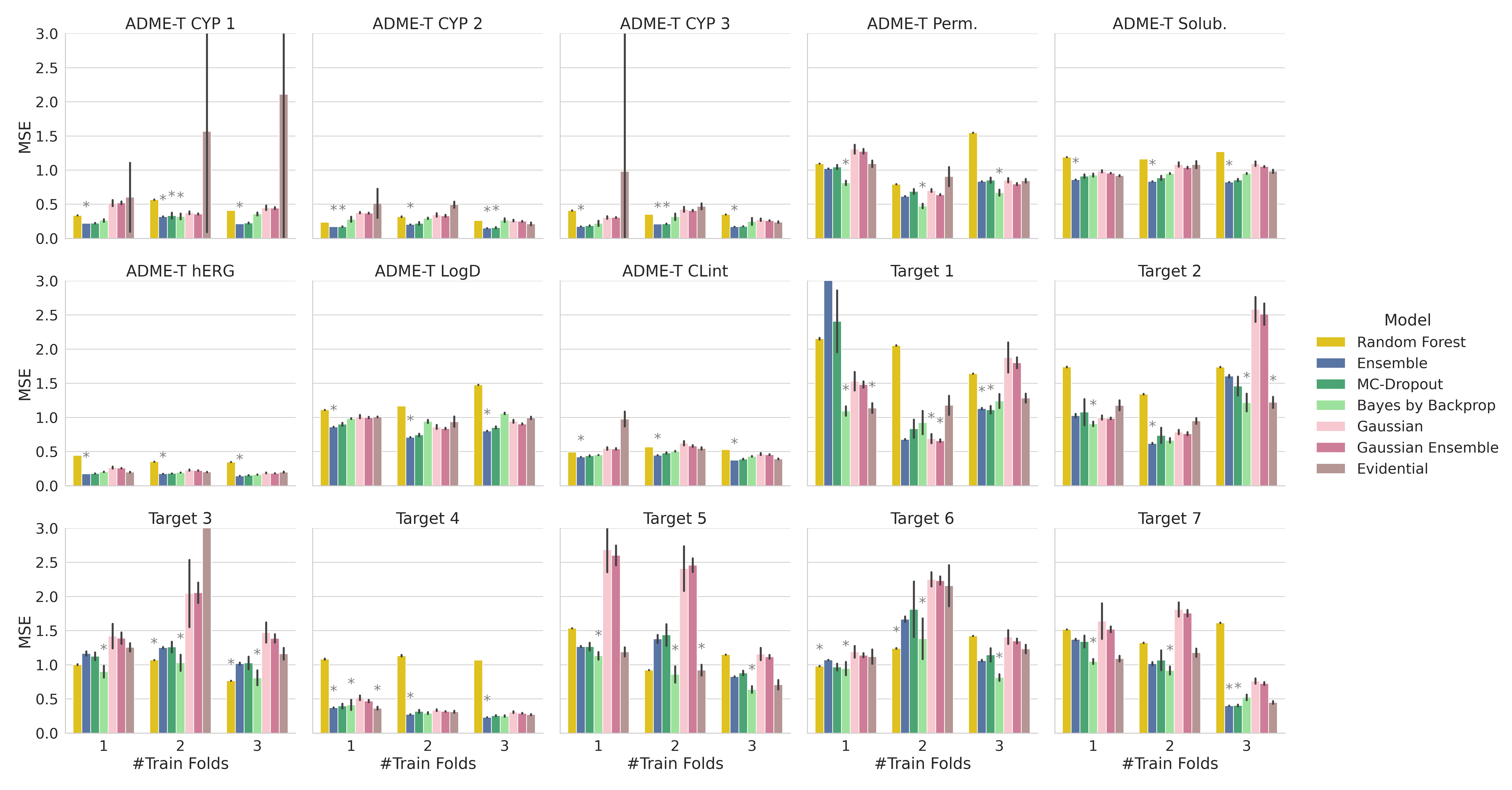}
            \caption{\textbf{Predictive Accuracy.} Comparing the predictive accuracy of all models in terms of MSE, aggregated over 10 experiments. For each dataset, the best model in terms of mean MSE is marked with a start together with any other models not statistically worse based on a one-sided Mann-Whitney-Wilcoxon test. Apart from the Random Forest and Evidential models, all other models are trained with censored labels.}
            \label{fig:mse}
        \end{figure}

        \paragraph{Predictive Accuracy.}
        The predictive performances of the models are presented in Fig.~\ref{fig:mse} in terms of MSE. An initial observation is that the MSE is generally lower for the models trained on the ADME-T assays. This result is expected given the diversity and size of the ADME-T assays, compared to the target-based assays. More data means that the models can learn to generalize better and the lack of shifts in both the feature space and the label space provides an easier prediction task. For seven out of the eight ADME-T assays, the Ensemble model is the best or among the best for all temporal settings. In a couple of instances, the MC-Dropout model is comparable, and for the ADME-T Perm. assay the Bayes by Backprop model significantly outperforms all other models. In the target-based assays, there is a greater variation in the best-performing models. For three target-based assays the Bayes by Backprop model performs the best across the temporal settings and in one target-based assay, the Ensemble model is the best. However, for three assays Target 1, Target 2, and Target 7 the best-performing model changes drastically over time. This observation is of great importance, as it illustrates that, typically, the conclusions about the best-performing model drawn early in the development of an ADME-T assay are reliable and still hold throughout the later development of the assay. On the other hand, as target-based assays might significantly shift in chemical space over time the same is not always true for the model comparison on these kinds of assays. For target-based assays, a new model comparison might be needed later on the the development of the assays to keep the best-performing model up-to-date. 

        For the most part, the Random Forest and Evidential models, which are trained without censored labels, perform poorly compared to the best-performing models. This is also true for the Evidential model on the two assays that do not have any censored labels, meaning that neither of the other models has the advantage of being trained on more data. We interpret this as an indication that the poor performance of the Evidential model compared to the other models is not primarily related to the fact that the other models are trained with censored labels. The Random Forest model, however, does reach comparable results to the Bayes by Backprop model for some of the temporal settings on the two assays, but it is never significantly better than all other models on any datasets. As such, it might be interesting to try to extend the Random Forest model to also handle censored labels in the future. However, in our analysis, it does not significantly outperform any of the other proposed methods herein. 

        Considering all datasets, i.e. assays and their respective temporal settings, the Ensemble model performs best overall, by being among the best-performing models for 27 out of the 45 dataset instances. The only approach that matches the performance of the Ensemble model is the Bayes by Backprop model which is among the best for 19 dataset instances. From the remaining models, the MC-Dropout model is only among the best-performing models for 6 dataset instances, together with the Ensemble model. The Evidential model and the Random Forest model are among the best together with the Bayes by Backprop model in 4 cases each. When training with two folds of the Target 1 assay,  the two Gaussian models outperform all other approaches significantly. However, this is the only instance where these models perform best. 
        
        It is reasonable and compatible with previous conclusions, that the Ensemble model achieves the highest predictive accuracy, i.e. lowest MSE \citep{ovadia2019can}. The reason is that this model contains the highest number of individually trained models. As such, the consensus in the predictions by the base models can reach a higher accuracy in cases where the problem is solvable by the available data. In comparison to the Gaussian Ensemble, the regular Ensemble has a slight advantage in terms of solely predictive accuracy because it is only trained to optimize the MSE. The Gaussian model simultaneously needs to optimize the calibration of the predicted uncertainty and thus might need to sacrifice some predictive accuracy if they are miss-aligned. For the Random Forest, the number of decision trees was optimized for each dataset in the model selection, as such in some cases the ensemble contains more than 50 trees. However, as the Random Forest is less complex than a neural network it is still reasonable that the neural network ensemble performs better. Out of the remaining models, the Evidential model relies on a single trained base model, but the Bayesian models instead learn an infinite set of model parameters.
        
        All in all, considering only the predictive accuracy of the considered models, our results show the Ensemble as the best-performing alternative. However, for practical applications, the computational cost of this model also needs to be taken into consideration. It is substantially more demanding to train 50 individual models for the Ensemble compared to the single model architecture used in the Bayesian approaches. As such, it is worth keeping in mind, that when Bayes by Backprop performs comparable to the Ensemble, it will also be significantly faster to train.

        \begin{figure}[t]
            \centering
            \includegraphics[width=\textwidth]{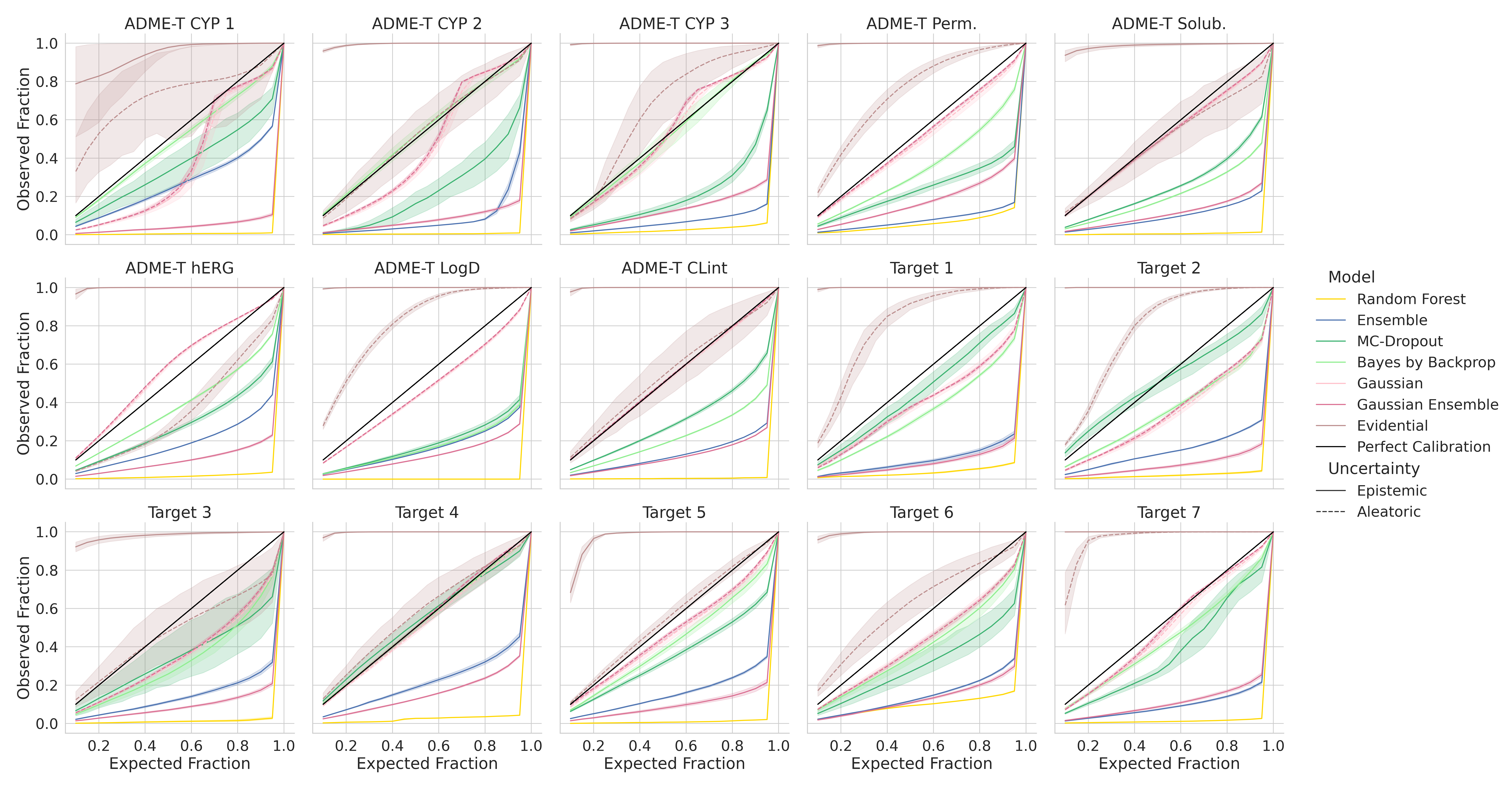}
            \caption{\textbf{Calibration Curves.} Full calibration curves for all uncertainty estimates on the third temporal setting containing three folds in the training set, aggregated over 10 experiments. The black line in each panel illustrates what a perfectly calibrated model would look like. Apart from the Random Forest and Evidential models, all other models are trained with censored labels.}
            \label{fig:calibration_curves3}
        \end{figure}

        \paragraph{Calibration.}
        Next, we evaluate the local calibration of the uncertainty estimates of each model in terms of the confidence-based calibration curves described in Section \ref{sec:methods} shown in Fig.~\ref{fig:calibration_curves3}. We focus this analysis on the third and biggest temporal setting. The results from the two earlier settings are provided in Fig.~\ref{fig:calibration_curves12} in Appendix \ref{app:model_comparison}. For the most part, we see that all aleatoric estimates are much better calibrated than many of the epistemic estimates. This is an interesting observation as it holds for models that produce both epistemic and aleatoric uncertainty estimates including the Gaussian Ensemble and the Evidential model, as well as the simple Gaussian model which only produces aleatoric estimates. The observation is true for ADME-T and target-based assays alike and thus irrespective of distribution shifts or dataset size. 

        Among the epistemic uncertainty estimates, a clear trend shows that the Evidential model is grossly under-confident, whereas many of the ensemble-based models are over-confident, especially the Random Forest model. This means that the Evidential model predicts CIs that are far too wide and, on the contrary, that the ensemble-based models predict very narrow CIs. For the three ADME-T CYP assays, the Bayes by Backprop estimates of the epistemic uncertainty are well-calibrated. Similarly, for many of the target-based assays, both the Bayes by Backprop model and the MC-Dropout model produce well-calibrated estimates of the epistemic uncertainty. These observations corroborate previously reported findings that ensembles tend to be under-calibrated in terms of the epistemic uncertainty compared to Bayesian approaches \citep{hubschneider2019calibrating}. 

        \begin{figure}[t]
            \centering
            \subfloat{\includegraphics[width=\textwidth]{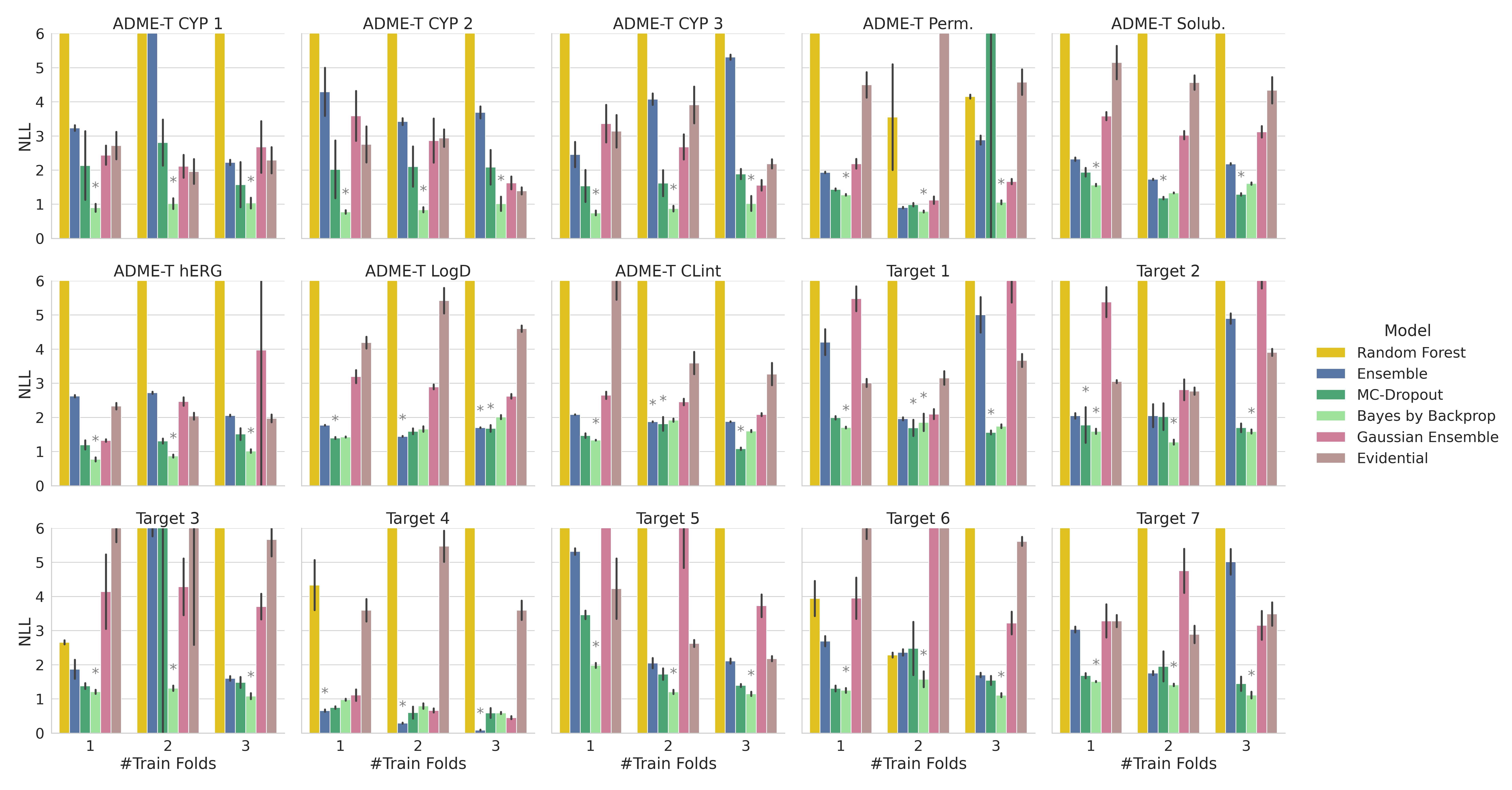}}
            \caption{\textbf{Combined Accuracy of Uncertainty Estimation and Predictive Performance.} Comparing the NLL of all epistemic uncertainty estimating models, aggregated over 10 experiments. For each dataset, the best model in terms of average NLL is marked with a star together with any other models not statistically worse based on a one-sided Mann-Whitney-Wilcoxon test. Apart from the Random Forest and Evidential models, all other models are trained with censored labels.}
            \label{fig:nll_epistemic}
        \end{figure}

        \paragraph{Overall Performance.}
        In the final step of our model comparison, we compare the models using the global scores NLL and ENCE which take both the predictive accuracy and the calibration of uncertainties into account. The results for NLL of the epistemic estimates are shown in Fig.~\ref{fig:nll_epistemic} while the results for NLL of the aleatoric estimates are provided in Fig.~\ref{fig:nll_aleatoric} of Appendix \ref{app:model_comparison}. Additionally, the ENCE scores are provided in Fig.~\ref{fig:ence} of the same Appendix, where similar conclusions can be observed as for the NLL scores. The NLL scores are highly varying for some of the poorly performing models, and therefore the plots have been cut to show the NLL below 2.5 in the case of the aleatoric models and 6 for the epistemic models. This was done for readability and does not hide any information about the best-performing models.  

        In terms of epistemic uncertainty, the Bayes by Backprop model is overwhelmingly best in terms of NLL, with 34 out of 45 best-performing scores. Only on the Target 4 assays, does the Ensemble model outperform Bayes by Backprop consistently throughout time. On 9 out of the 15 assays in total, the Bayes by Backprop model outperforms all other models entirely on all temporal settings. This observation means that in terms of the NLL, the conclusions drawn early for an assay from both categories can oftentimes be trusted to hold throughout time. We believe that this robustness for the target-based assays that exhibit distribution shifts speaks to the reliability of the epistemic uncertainty estimates in accounting for the distribution shifts where present. As such, the best-performing model in terms of predictive accuracy might change over time, only due to differences in how well the models handle the distribution shifts, but the overall best-performing model is reliable for all kinds of assays irrespective of distribution shifts.

        Similarly to what was observed in the MSE, we see in Fig.~\ref{fig:nll_epistemic} that the NLL scores are generally slightly lower for the ADME-T assays compared to the target-based assays. Target 4 is again an outlier with quite low NLL scores compared to the other target-based assays. Recall from eq. \ref{eq:nll} that the NLL includes a term with the squared error. Thus, it is likely that this trend arises from that source. In the ENCE scores presented in Fig.~\ref{fig:ence} in Appendix \ref{app:model_comparison}, the same trend can be partially observed but it is not as prominent. 

        For the aleatoric uncertainty estimates, shown in Fig.~\ref{fig:nll_aleatoric} in Appendix \ref{app:model_comparison}, the Evidential model is significantly outperformed by either or both of the Gaussian models in all but two of the 24 ADME-T dataset instances. Among the target-based assays, there are more times that the Evidential model outperforms the Gaussian models. In particular, for the Target 5 assay, the Evidential model is predominately better. However, in the vast majority of cases among all datasets, 39 out of 45, the Gaussian Ensemble is the best or among the best, sometimes also beating the single Gaussian model. In the case of the NLL for the aleatoric uncertainty estimates, like for the MSE, the best-performing model is almost exclusively robust throughout time for the ADME-T assays but not always robust for the target-based assays. Note also that the NLL scores for the aleatoric uncertainty estimates are mostly lower than for the epistemic uncertainty estimates, similar to the results seen in the calibration curves. 

    \subsection{Case Study.}
        Finally, we perform a case study on one example assay from each category to explore the practical implications and usefulness of the epistemic and aleatoric estimates respectively. In the case study of aleatoric uncertainty, we compare the predicted uncertainty estimates with ground truth experimental error in the form of the standard deviation between duplicated experiments. For this analysis, we cannot include censored labels due to the aggregation strategy described in Section \ref{sec:methods}. Additionally, we exclude experiments repeated less than three times and those with no variation reported in the data. We chose the assays with the most available compounds with duplicated experiments, e.g. ADME-T hERG and Target 6. For the epistemic case study, we instead pick Target 7 as a target-based assay due to its particularly challenging distribution shift. Similarly to the aleatoric case study, we compare the results with the ADME-T hERG assay, which has no visible distribution shift in either the feature or label space as seen in Fig.~\ref{fig:label_distributions} and Fig.~\ref{fig:tsne_distributions} in Appendix \ref{app:distributions}.

        This case study is solely possible due to the temporal split and the additional information available through the internal pharmaceutical data, which are generally not accessible or trustworthy in public datasets. Similar remarks were made by \citet{sheridan2013time} and point to the necessity of openness from the pharmaceutical industry to learn from each other and push the field forward. 

\begin{figure}[t]
            \centering
            \includegraphics[width=\textwidth]{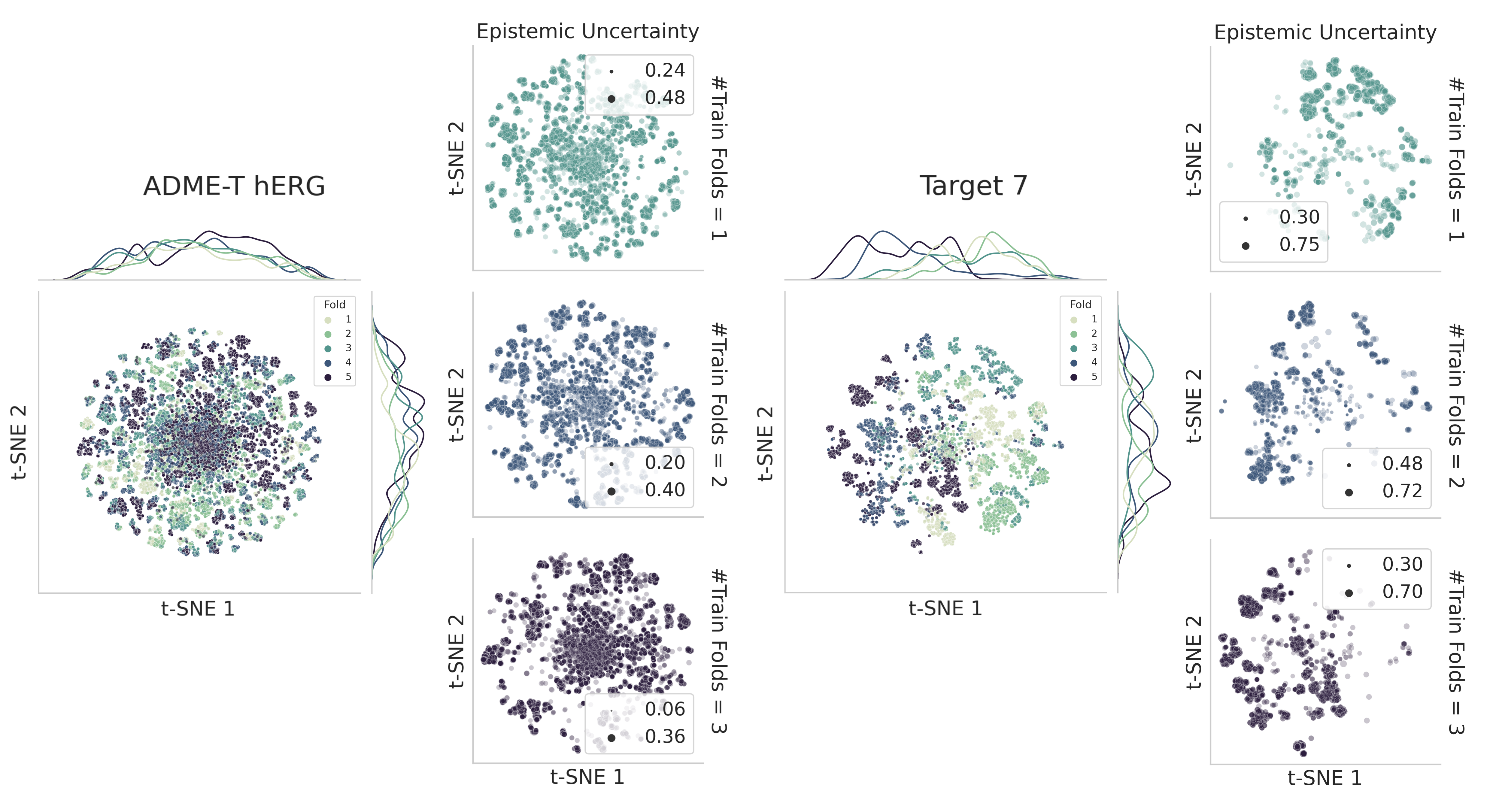}
            \caption{\textbf{Epistemic Estimates by Bayes by Backprop.} A practical illustration of the epistemic uncertainty estimates by the Bayes by Backprop model on two assays: ADME-T hERG without clear distribution shifts over time, and Target 7 with prominent distribution shifts over time.}
            \label{fig:case_study_epistemic}
        \end{figure}

        \paragraph{Epistemic Uncertainty Quantification.}
        For the practical case study of the epistemic uncertainty estimates, we illustrate the epistemic uncertainty estimates averaged over the 10 repeated experiments of the Bayes by Backprop model on the ADME-T hERG and the Target 7 assays. The Bayes by Backprop model is chosen as it is significantly the best model in terms of NLL across all temporal settings of both of these example assays, as shown in Fig.~\ref{fig:nll_epistemic}. The case study compares the epistemic uncertainty between different regions of the t-SNE projections of each test set, as illustrated in the rightmost part of each panel in Fig.~\ref{fig:case_study_epistemic}. It also relates this to the full t-SNE projection including the folds used for training and validation, as seen in the leftmost parts of the same figure. Note, that this only accounts for distributional shifts in a linear approximation of the feature space. As the fundamental nature of the feature space is non-linear, this analysis cannot cover these non-linearities. Similarly, we cannot evaluate any parts of the epistemic uncertainty related to other factors than the distribution shift in this analysis. The other factors might include epistemic uncertainty that relates to the choice of model architecture or training procedures. We have to rely on the metrics used in the model comparison section above to account for these factors of epistemic uncertainty.

        In the full t-SNE projections of each dataset containing all five folds, we see the trend described in Section \ref{sec:methods}. The chemical space of the ADME-T hERG assay is highly diverse, already in the first fold, and does not shift substantially over time. However, some clusters are formed in the outer edges of the chemical space, where the different folds are well separated. For the Target 7 assay on the other hand, there is a clear shift in the chemical space from the right side of the plot representing the initial measurements to the left side showing the later folds. Similar trends are also present in the label distributions as seen in Fig.~\ref{fig:label_distributions} in Appendix \ref{app:distributions}. Furthermore, the label distribution for the Target 7 assays does not shift continuously over time. Instead, it shifts greatly toward higher pIC50 values in the second fold and then back toward lower values in the last two folds. The remaining three plots to the right in each panel of Fig.~\ref{fig:case_study_epistemic} show the t-SNE projections of each test set, i.e. folds 3, 4, and 5, separately. Here, the size of the data points is determined by the averaged epistemic uncertainty predicted by the Bayes by Backprop model trained on the three temporal settings respectively, i.e. with an increasing number of training folds. Note that the legends of these plots detail the respective minimum and maximum predicted uncertainties on the given test set. 

        For the ADME-T hERG assay, only minor differences can be observed between compounds positioned in various regions of the feature space. This is expected given the lack of overall distribution shifts between the folds. Additionally, we evaluate the span of uncertainty estimates, according to the minimum and maximum values listed in the legend of each plot. The third temporal setting has the largest span of 0.3 compared to 0.24 and 0.2 for the first two settings. This indicates that the model is better at separating the compounds in the final temporal setting compared to the previous one, likely due to the increased size of the training data. Another effect of the increasing training data can be seen in the difference in the maximum predicted uncertainty between the folds. For the model trained on three folds, the maximum is 0.36, which is smaller than the values seen for the models trained on less data, which are 0.4 for the model trained on two folds and 0.48 for the model trained on one fold. This is a direct and expected behavior of the epistemic uncertainty, that more data results in more certain models i.e. lower predicted uncertainties. In the case of the target-based assay, we note that the maximum estimated uncertainty on each test set is very high, between 0.7-0.75. This can be a result of the general distribution shift present between every fold or the overall small dataset. In both cases overall high uncertainty is to be expected.
    
        Based on the distribution shift present in the feature space, and considering that epistemic uncertainty should account for distributional shifts, we expect that distribution shifts are reflected in the predicted epistemic uncertainties. As such, our analysis provides empirical evidence to support this claim, and it illustrates that the uncertainty estimates cover additional sources of uncertainty related to the model itself, such as poor generalization in cases when training data is limited. It is important to understand all sources of uncertainty when basing future critical decisions on model predictions, such as in drug discovery. 
        
        Considering the cases addressed in this section of our study, we can deduce practical suggestions on how the identified sources can impact the continued drug discovery process. If poor generalization is determined, indicated by overall high uncertainty estimates and low predictive performance of a model, the model needs to be retrained with more data before deployment. Another alternative is to reconsider the choice of model, but our temporal study shows that the Bayes by Backprop model remains the best choice in the future, despite the addition of more data. When distribution shifts are identified, such as in the target-based assay, further data exploration in chemical spaces with high uncertainty estimates is needed before deployment. 
    
        Further research is necessary to disentangle the sources of epistemic uncertainty, including distribution shifts and other model-related sources. One alternative approach would be to quantify the distribution shift using other means, either with distance-based approaches, such as the average Tanimoto or cosine similarity \citep{sheridan2004similarity} between an inference compound and compounds in the training set, or the interpretable method proposed by \citet{kulinski2023towards}. Additionally, more advanced pre-training procedures can be used, that are trained to incorporate distribution shifts more effectively \citep{bertolini2023explaining}. After the distribution shift has been independently quantified, the predicted epistemic uncertainty could be re-evaluated such that the remaining model uncertainty is disentangled from this information.

        \paragraph{Aleatoric Uncertainty Quantification.}
        To examine the aleatoric estimates by the best-performing model from the model comparison above, we need a way of quantifying the true experimental error. As such, we consider the variation between duplicated measurements of a single compound on the assay to provide this information. However, note that this information was not provided to the models during training because duplicated measurements were aggregated as part of the data preprocessing described in Section \ref{sec:methods}. This procedure of aggregation is common when applying machine learning to molecular property prediction, but an important question is if the available methods to estimate aleatoric uncertainty can learn the experimental error from such preprocessed data.
        
        One potential issue with the described estimates of the true experimental error is that the number of duplicated measurements varies greatly between compounds. Some compounds are tested many times whereas many are tested only a handful of times or only once. The compounds for which there is only one measurement do not give any indication about the experimental error and ones where only a few duplicates are available might not be as reliable as those where the experiment was repeated many times. For this reason, we consider only compounds from the test sets that have more than 2 measurements and ones where the variation was greater than exactly zero. 
        
\begin{figure}[t]
            \centering
            \subfloat{\includegraphics[width=0.4\linewidth]{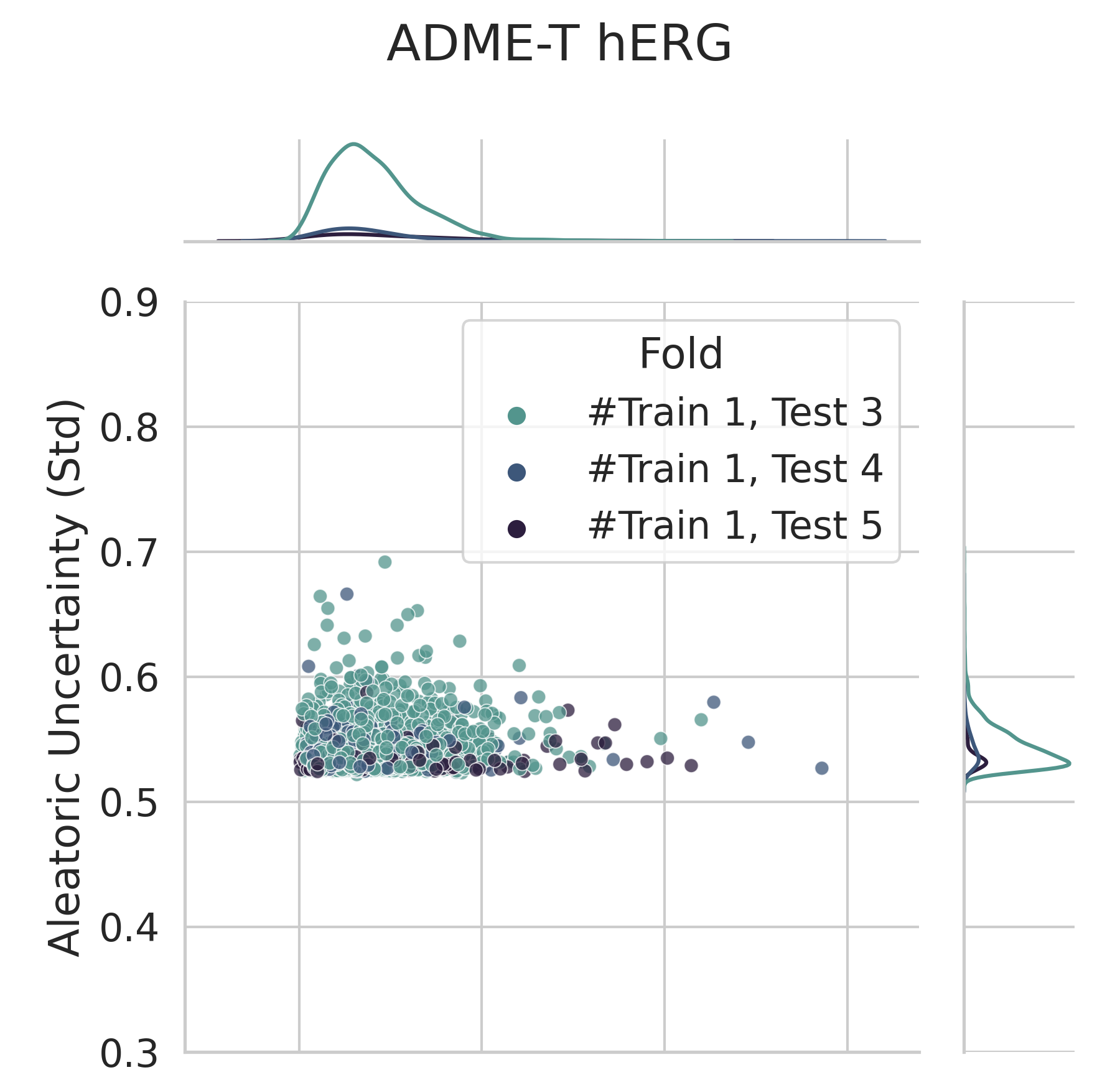}}
            \subfloat{\includegraphics[width=0.4\linewidth]{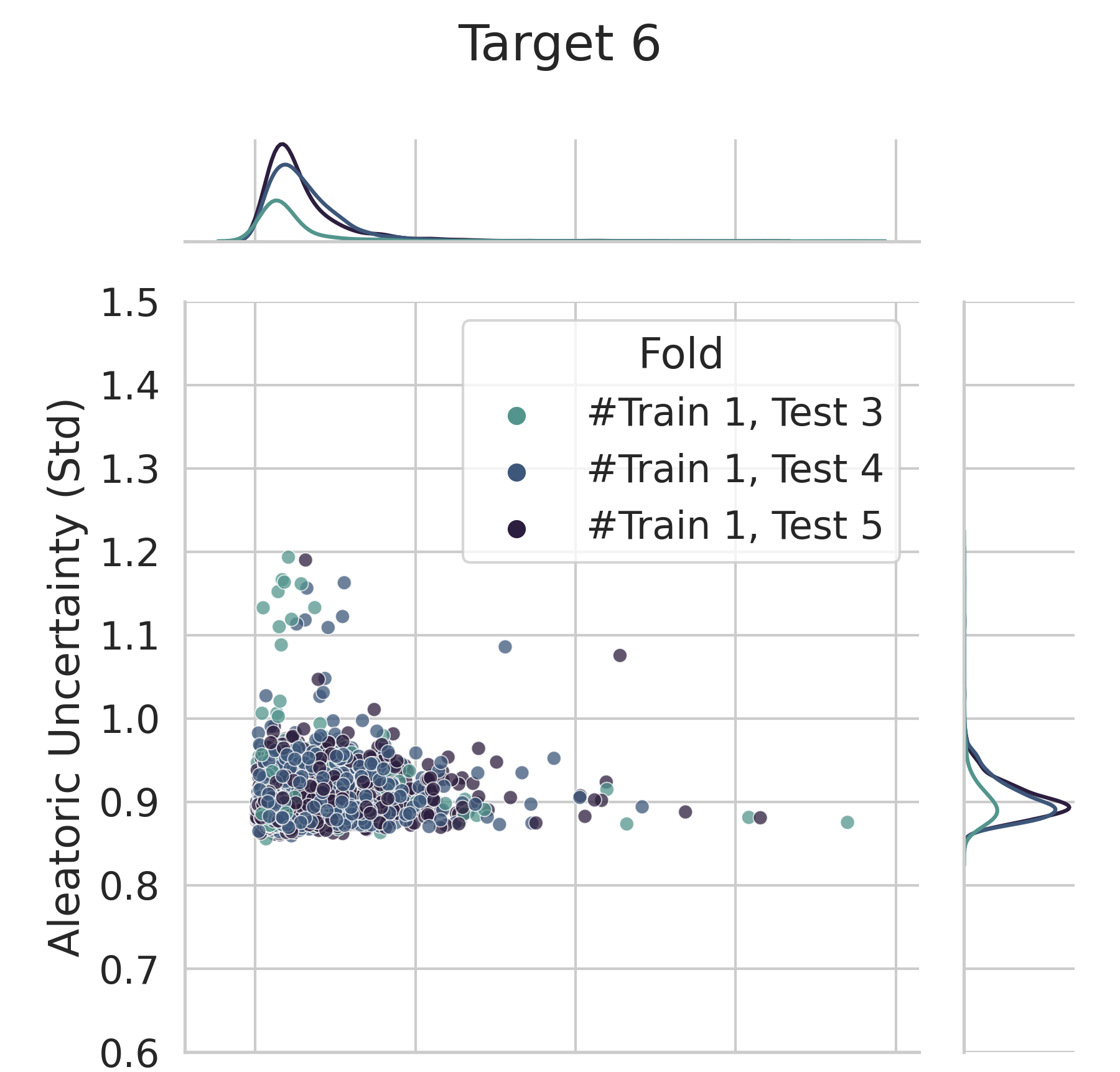}} \\
            \subfloat{\includegraphics[width=0.4\linewidth]{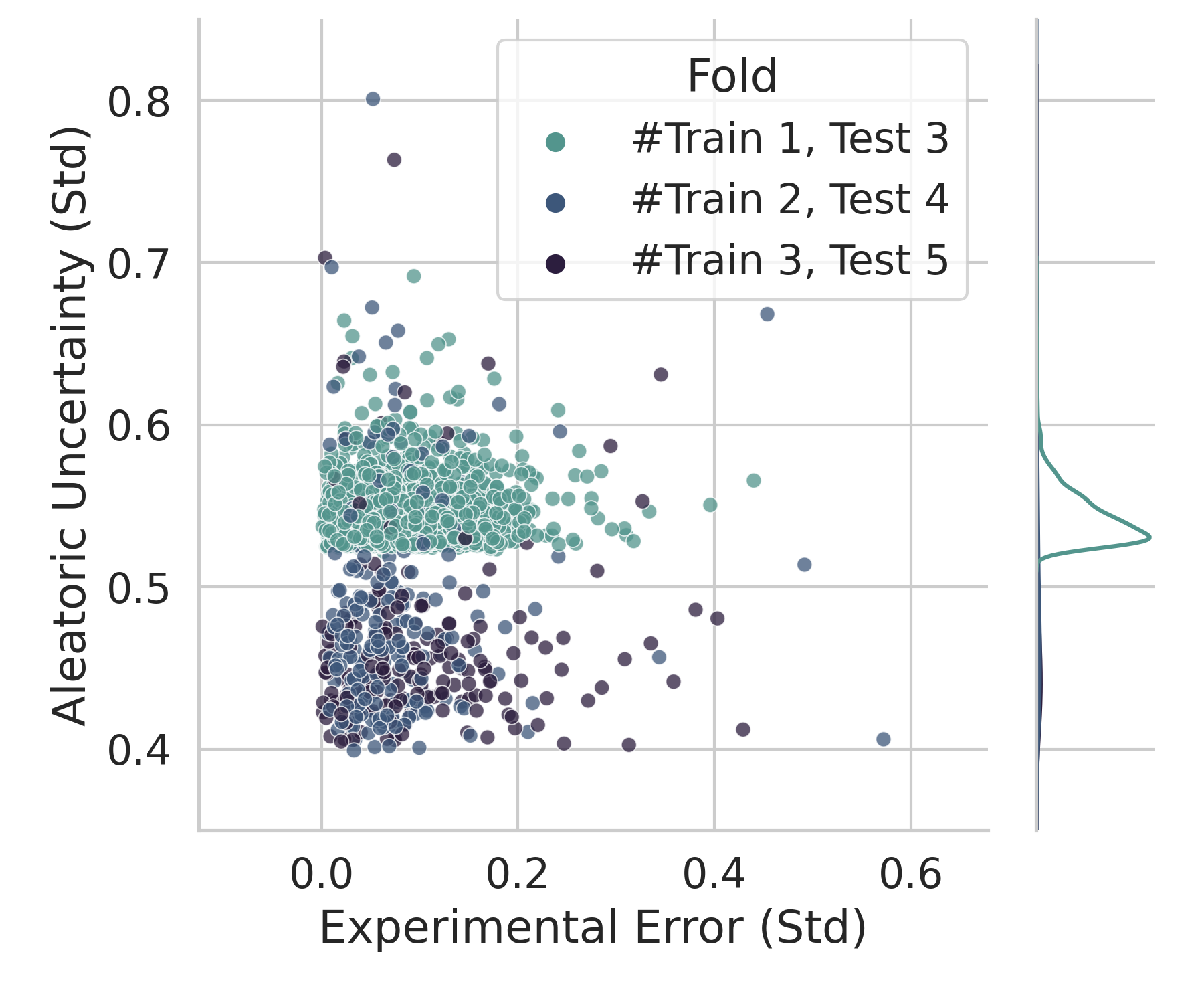}}
            \subfloat{\includegraphics[width=0.4\linewidth]{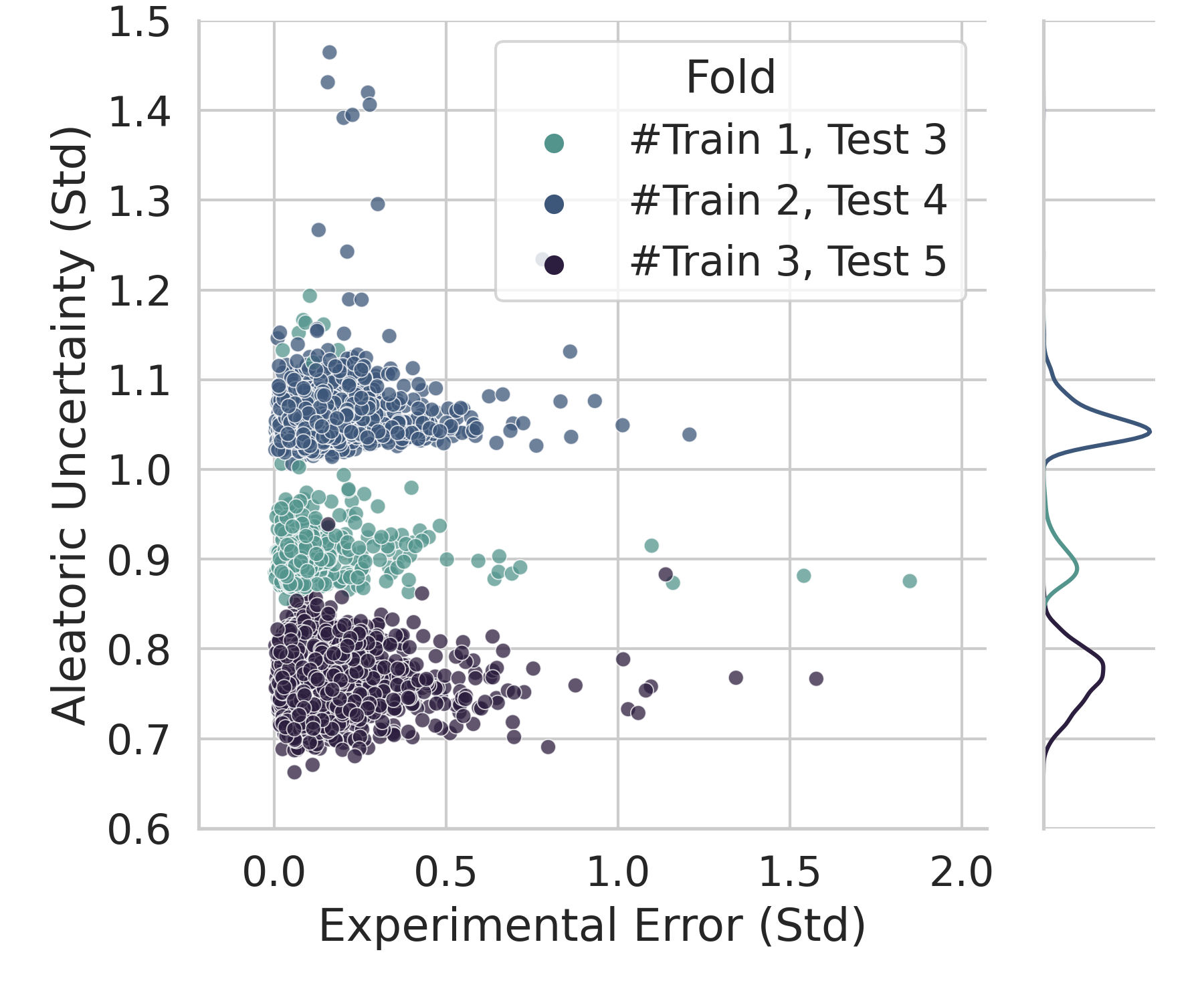}} \\
            \caption{\textbf{Aleatoric Estimates by Gaussian Ensemble.} A practical illustration of the aleatoric uncertainty estimates by the Gaussian Ensemble on the two assays with the most available duplicate experiments in the test sets. Comparing the predicted aleatoric uncertainty to the standard deviation of duplicated experiments, i.e. experimental error. The top row shows prediction by only the model trained on the first fold. The bottom row shows each test set predicted by the respective, different models trained on all folds until the given test set.}
            \label{fig:case_study_aleatoric}
        \end{figure}

        After filtering for duplicated experiments in the available data, the ADME-T hERG assay contains 712 compounds with duplicated measurements from the test sets of all three temporal settings combined. The Target 6 assay similarly contains 1712 duplicated experiments in total between the three test sets. In this analysis, we focus on the aleatoric estimates averaged over the 10 repeated experiments of the Gaussian Ensemble. In the previous experiments of this study, this approach emerged as the significantly best-performing model across all temporal settings of the ADME-T hERG assays and as one of the highest-ranked approaches on the Target 6 assay. Fig.~\ref{fig:case_study_aleatoric} illustrates the relationship between the predicted aleatoric estimates and the available experimental errors, colored by the three test sets. 

        In general, we cannot necessarily expect the model to learn the correct scale of the experimental errors since the experimental error is not included in the training data. Instead, we are looking to evaluate if any upward trends can be observed, such that compounds with larger experimental error also prompt greater predicted aleatoric uncertainty. However, no such trend can be observed for any of the models or test sets on the two assays. Given that the aleatoric estimates are still well-calibrated according to the observed errors, as seen in the calibration-curves of Fig.~\ref{fig:calibration_curves3} and the overall scores shown in Appendix \ref{app:model_comparison}, this could indicate that the learned aleatoric uncertainties reflect something other than the experimental error examined here. The model may be able to learn trends in the label space given similar compounds in the feature space. For instance, activity cliffs can exist where very similar compounds prompt significantly different responses in an assay \citep{maggiora2006outliers}. Such relationships could be explored in future research, but from our analysis, it would also be interesting to try modeling these molecular properties without aggregating duplicated measurements. 
        
        Additionally, we compare the scale of the predicted aleatoric estimates across different test sets and assay types. First, we see that the estimates are significantly lower on the ADME-T hERG assay than the Target 6 assay. This might relate to several factors that make the two assays distinctly different. It could simply be a result of the overall size of the datasets: recall from Table \ref{tab:data_overview} that the Target 6 assay has only 13,093 labels compared to the 67,687 labels available for ADME-T hERG. However, this trend also corresponds well with the observed standard deviation of the control compounds on the two assays, which was 0.19 for the ADME-T hERG assay and 0.31 for the Target 6 assay. These errors from the control compounds should be more reliable and thus give a better general estimate of the overall homoscedastic noise present in each assay.

        Furthermore, an important discrepancy can be observed by comparing the distributions of the aleatoric estimates on each test set between the first row and the second row of Fig.~\ref{fig:case_study_aleatoric}. In the first row, all predictions are made by the same model, the one trained on only the first fold. Here, the distributions of the aleatoric estimates are roughly the same for each test set of the two example assays respectively. In the second row, the illustrated predictions of each test set are made by different models, trained on an increasing amount of folds. Thus, these are the models used throughout all previous results of this study. Apart from these models being trained on different amounts of data, they are also optimized individually as explained in the model selection in Appendix \ref{app:model_selection}. Therefore, these models can have vastly different model architectures and be trained using different training procedures such as learning rate. 
        
        Crucially, none of these factors should have any effect on the aleatoric uncertainty as it is often categorized in literature as irreducible \citep{apostolakis1990concept, kendall2017uncertainties, hullermeier2021aleatoric}. Despite this, we do see distinct differences between most of the distributions of aleatoric estimates by each of these different models in the bottom row of Fig.~\ref{fig:case_study_aleatoric}. Specifically, we see some instances where increasing amounts of training data result in generally lower aleatoric estimates. However, there is also one case where the opposite is true, between test folds 3 and 4 of the Target 6 assay. We can only assume this relation to be because of the differences in model architectures. Therefore, our empirical results raise questions about whether the explored models' supposed estimates of aleatoric uncertainty are really fully disentangled from the epistemic uncertainty. Similarly, recent theoretical work on deriving suitable measures for aleatoric, epistemic, and total uncertainty has found that the aleatoric and epistemic parts do not necessarily have to add up to the total uncertainty \citep{sale2024secondorder}. As such, the disentanglement of the sources of uncertainty should be considered an ongoing field of research that needs more work to fully determine how these estimates should be categorized and how they relate to the underlying noise in the data.

\section{Conclusions}

    The low-data challenge in drug discovery is typically accompanied by additional, often overlooked, partial information from censored labels. Despite their potential value, censored labels have not yet been fully utilized due to the lack of suitable methods to incorporate them during machine learning. Recognizing this gap, we have developed extensions to several established machine learning models, enabling them to effectively learn from censored labels while providing robust and reliable uncertainty quantification. Our results showed a particular advantage of including the censored labels for ensemble-based and Bayesian models but also enhanced Gaussian models in cases where >35\% of the data contained censoring. 
    
    Through a comprehensive temporal study using internal pharmaceutical datasets, we demonstrated the importance of these extended models in accurately predicting key affinity scores, and side effects, of potential drug compounds, and their associated uncertainties. Our model comparison included approaches that estimate both aleatoric and epistemic uncertainties, providing a more complete understanding of prediction confidence. Specifically, we found that a straightforward ensemble of individually trained neural networks achieves generally high predictive accuracy. However, when also accounting for the calibration of uncertainty and computational cost, we recommend using Bayes by Backprop.

    Thanks to the temporal evaluation, we were able to detect key differences between the distributions of target-based assays versus ADME-T assays. The ADME-T assays are more diverse in terms of the chemical space and thus exhibit less of a shift throughout time compared to the target-based assays where distinct differences could be observed for different time points. In light of these trends, we showed that the best-performing model in terms of predictive accuracy and calibration of aleatoric uncertainty are typically robust throughout time for ADME-T assays but not always for target-based assays. As such, an evaluation of the best model for ADME-T assays can be trusted to hold without reevaluation whereas target-based assays may need to be reassessed occasionally. 
    
    Finally, our case study showed how the uncertainty estimates from the most effective models can be practically applied to inform and guide ongoing drug development efforts, offering valuable insights for risk management and decision-making. Herein, we found that the epistemic uncertainty estimates correlate well with theoretical assumptions, but that the aleatoric uncertainty estimates require further analysis to understand their relation to the underlying, inherent noise in the data.

\section*{Acknowledgements}
    We thank our colleagues for their valuable feedback. This study was partially funded by the European Union’s Horizon 2020 research and innovation programme under the Marie Skłodowska-Curie Actions grant agreement “Advanced machine learning for Innovative Drug Discovery (AIDD)” No. 956832. HRF, and AA are affiliated to Leuven.AI and received funding from the Flemish Government (AI Research Program).

\section*{Data Availability Statement}
    This work was conducted on internal data that cannot be disclosed beyond the information provided in Table \ref{tab:data_overview}. Despite this, we believe the work brings great value to the community by showcasing crucial aspects of the data that could not otherwise be analyzed. First, the temporal evaluation would not be possible to the same extent on public data, which led to conclusions about how the models compare over time. Second, while censoring in experimental labels are naturally occurring in internal data, it is less common in public data. Therefore, the key contribution of adapting and evaluating current uncertainty quantification approaches to censored labels, would not be possible without the proprietary data. All methodology will however be made available in our code, soon to be published on GitHub under the MIT license. Instructions to prepare the programming environment, as well as how to run the training, inference, and evaluation procedures on similar public data from Therapeutics Data Commons \citep{huang2021therapeutics} can also be found there. Throughout this work, all experiments were run on a cluster of servers with diverse Nvidia GPUs using Python 3.11 with PyTorch 2.0.1 \citep{paszke2019pytorch}.

\section*{Disclosure of Interests} 
    The authors have no competing interests to declare that are relevant to the content of this article.

\bibliographystyle{unsrtnat}
\bibliography{references}

\newpage
\appendix

\section{Temporal Distributions}
\label{app:distributions}
\begin{figure}[ht]
        \centering
        \includegraphics[width=\textwidth]{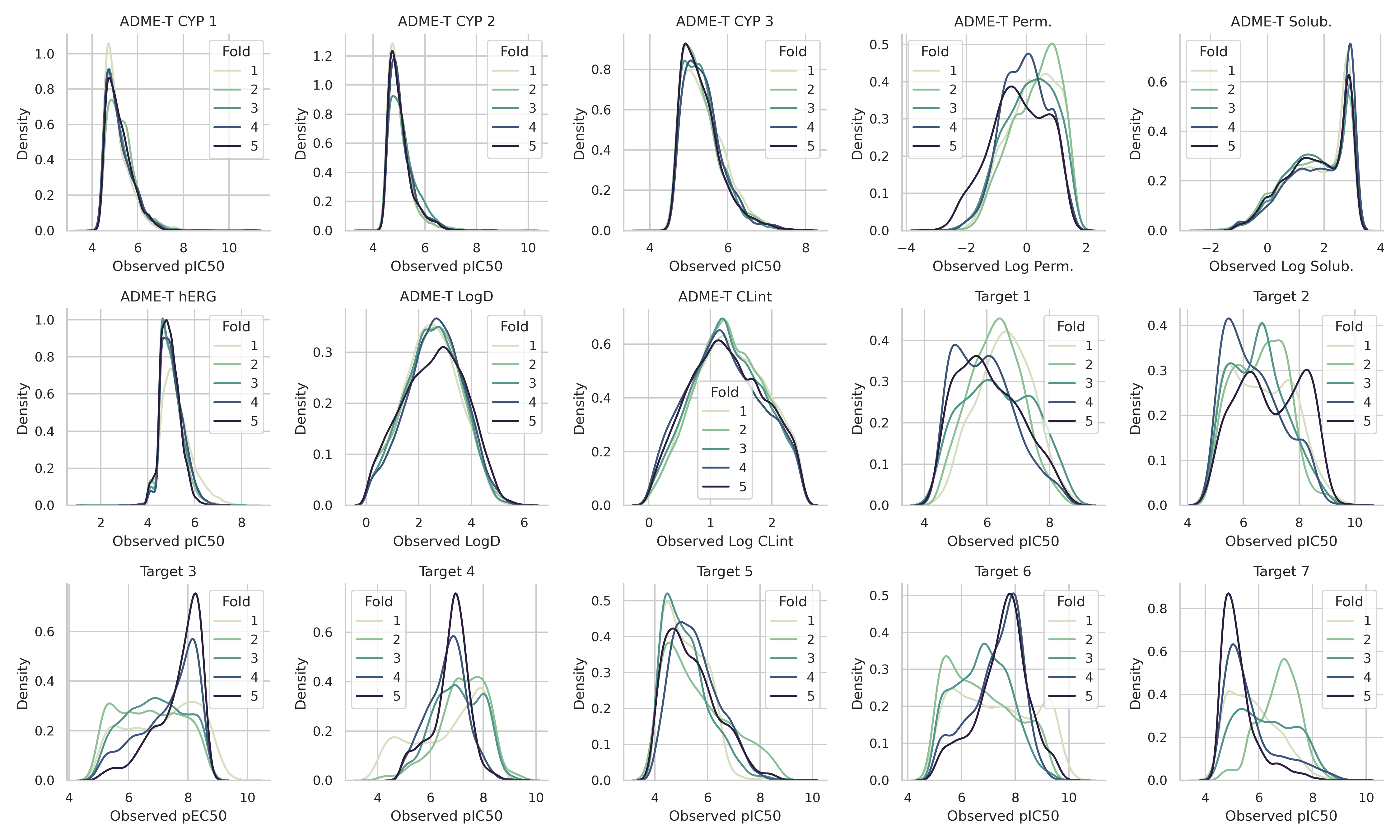}
        \caption{\textbf{Temporal Distribution of the Label-space.} Distribution of observed experimental labels across the five temporal folds illustrated per assay.}
        \label{fig:label_distributions}
    \end{figure}

\begin{figure}[ht]
        \centering
        \subfloat{\includegraphics[width=0.2\textwidth]{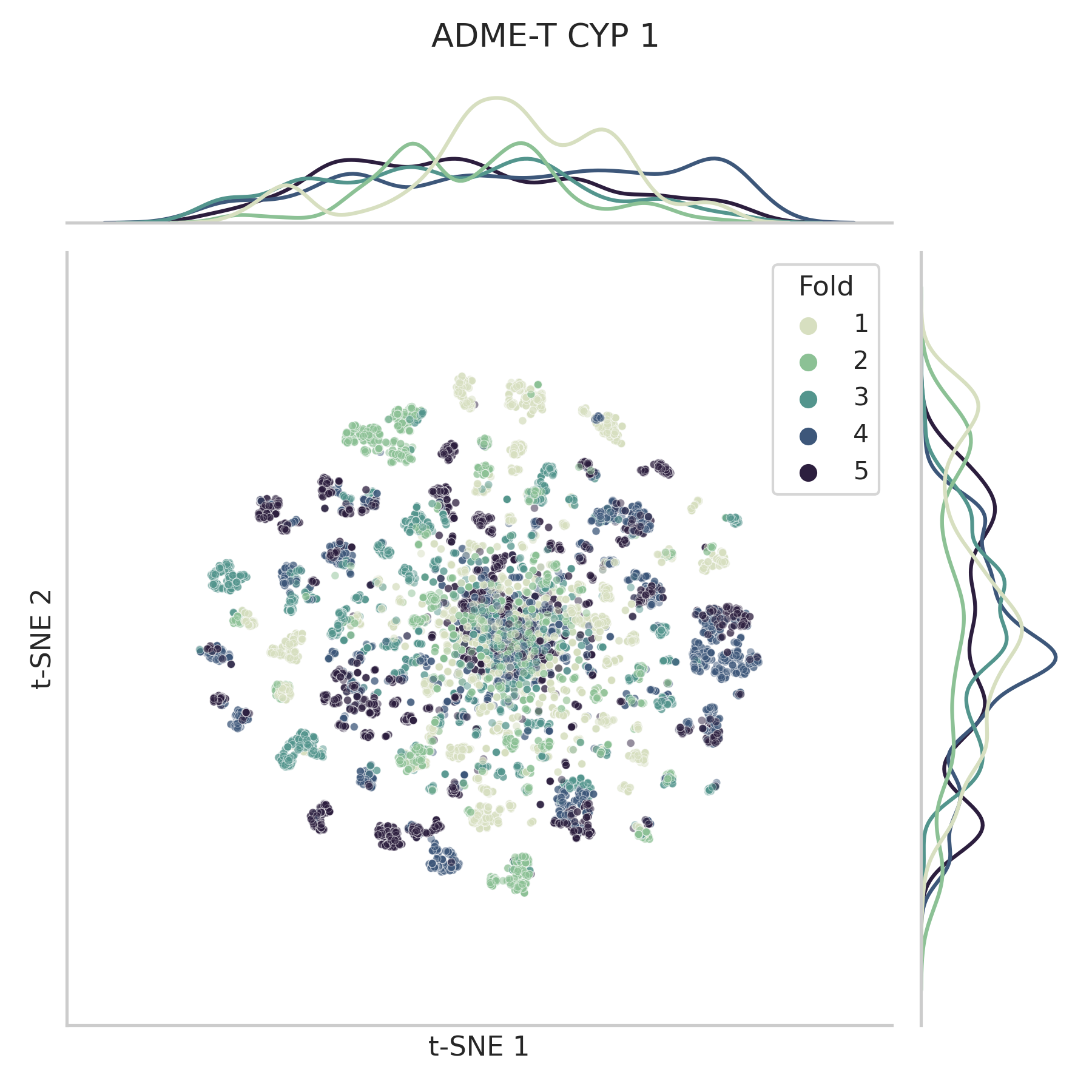}}
        \subfloat{\includegraphics[width=0.2\textwidth]{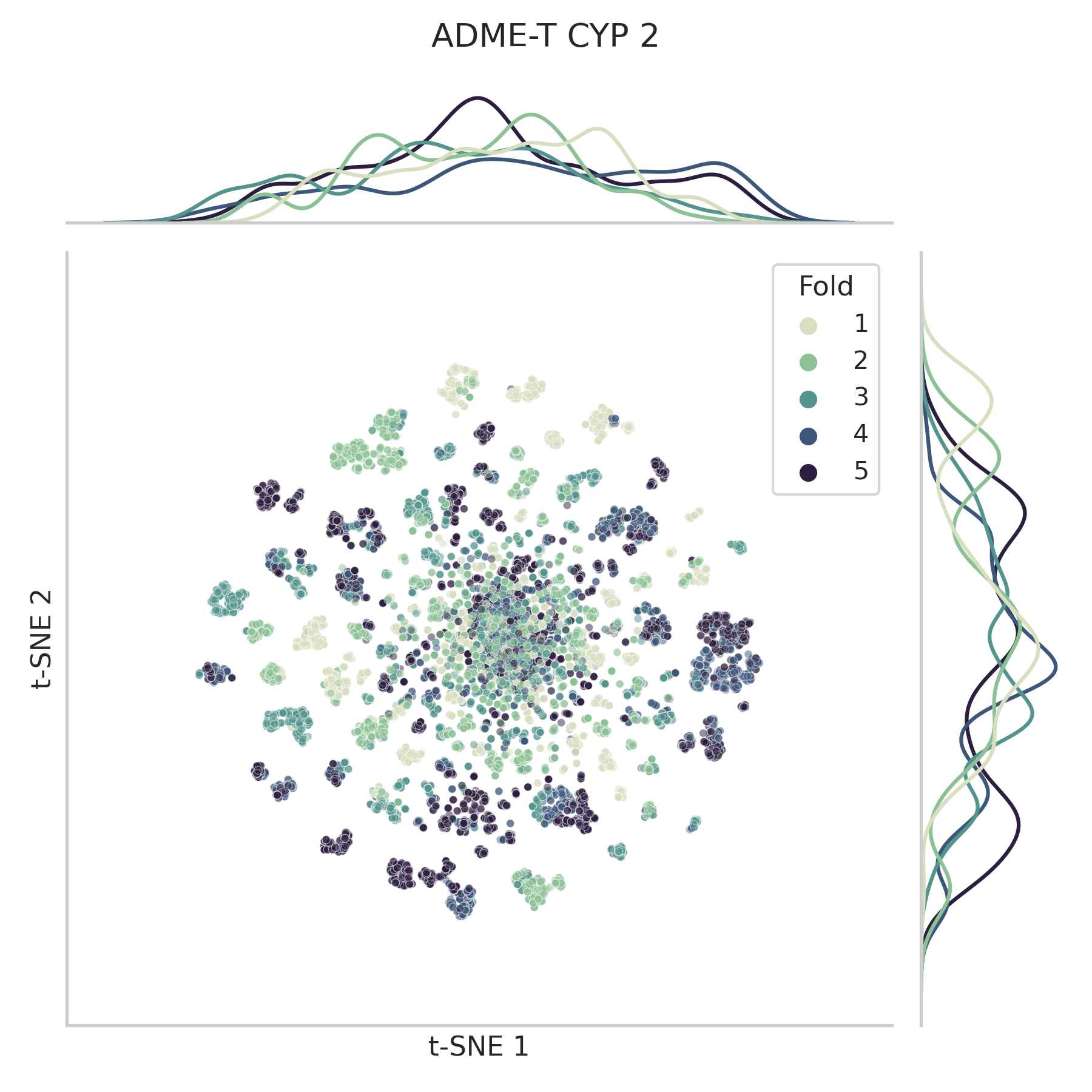}}
        \subfloat{\includegraphics[width=0.2\textwidth]{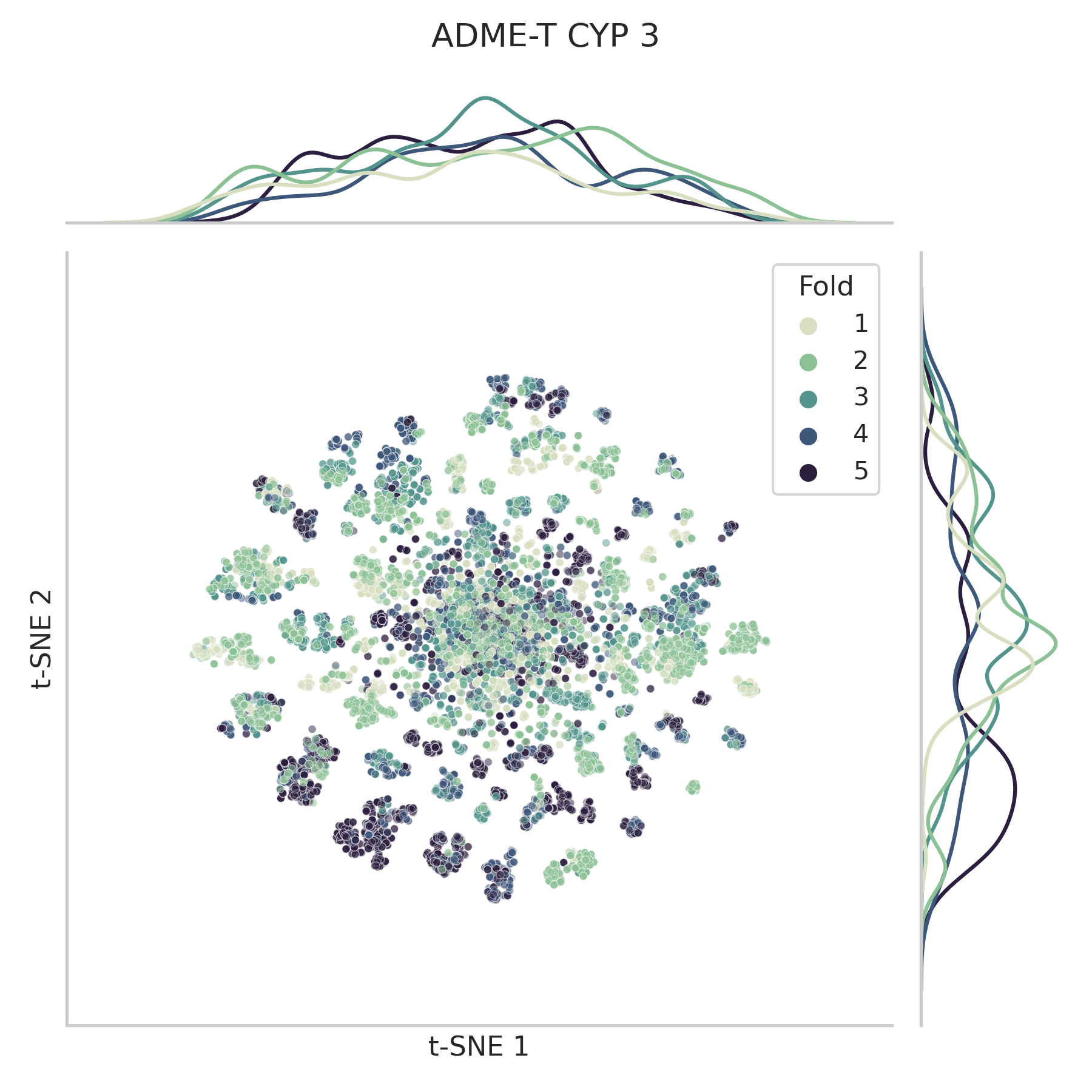}} 
        \subfloat{\includegraphics[width=0.2\textwidth]{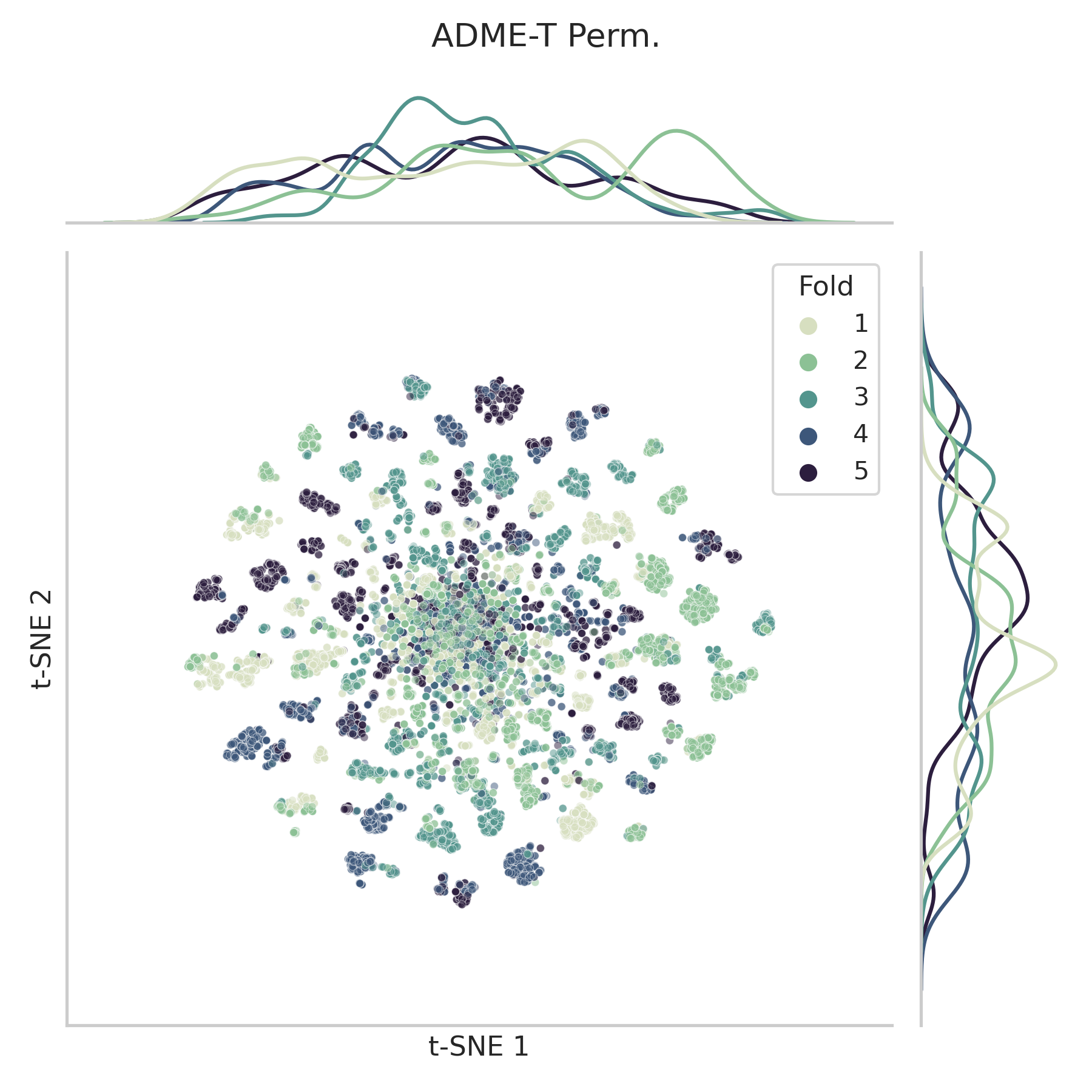}}
        \subfloat{\includegraphics[width=0.2\textwidth]{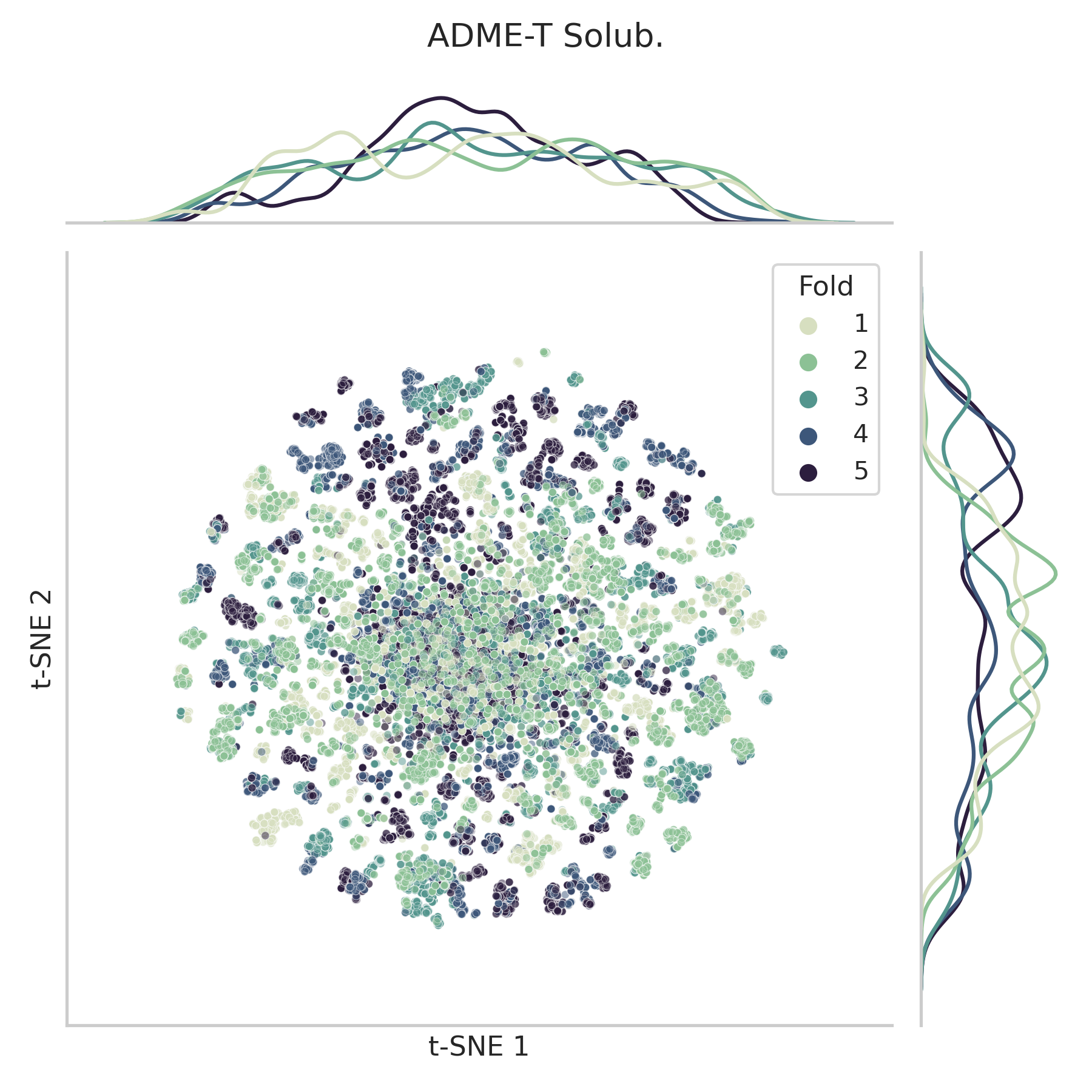}} \\
        \subfloat{\includegraphics[width=0.2\textwidth]{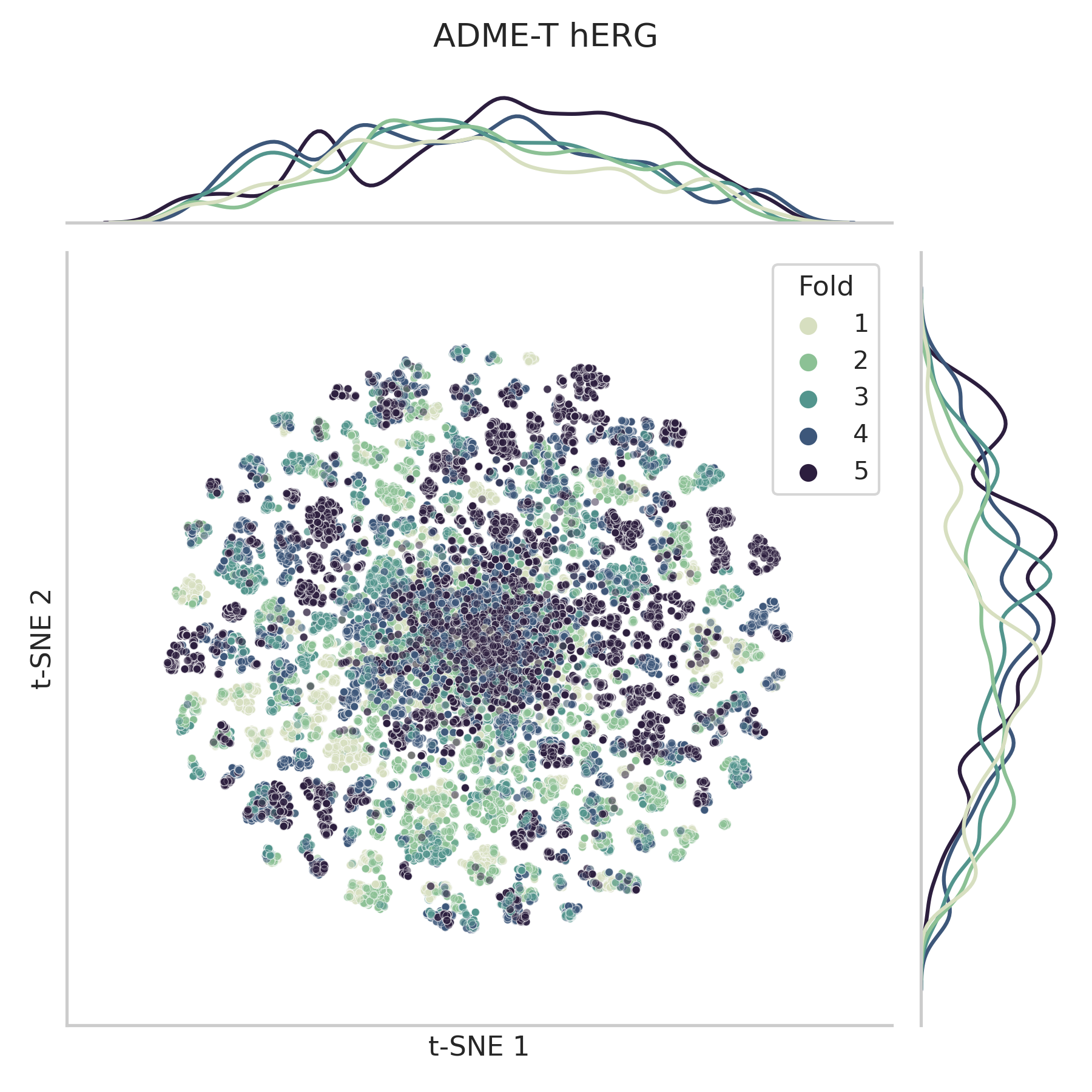}}
        \subfloat{\includegraphics[width=0.2\textwidth]{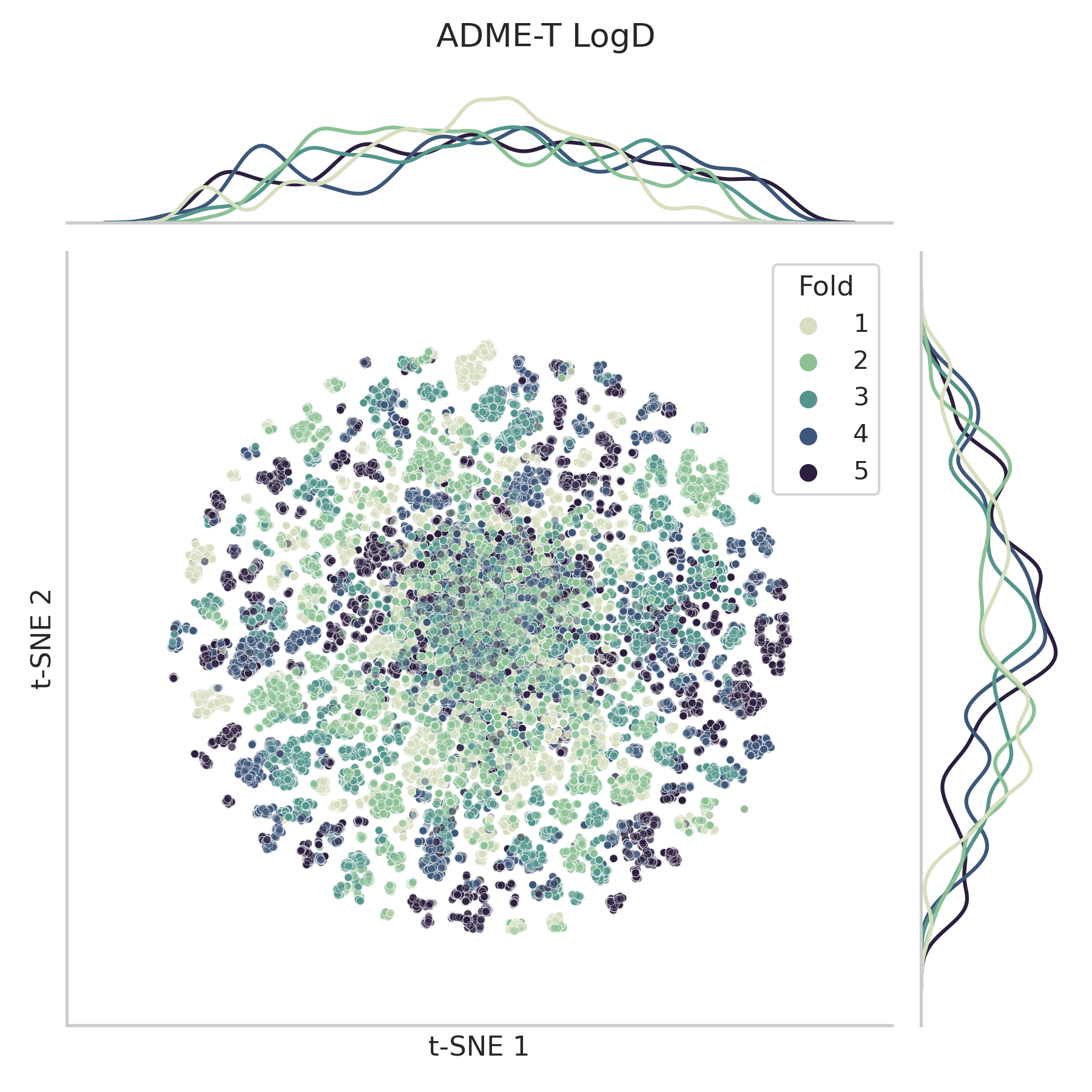}}
        \subfloat{\includegraphics[width=0.2\textwidth]{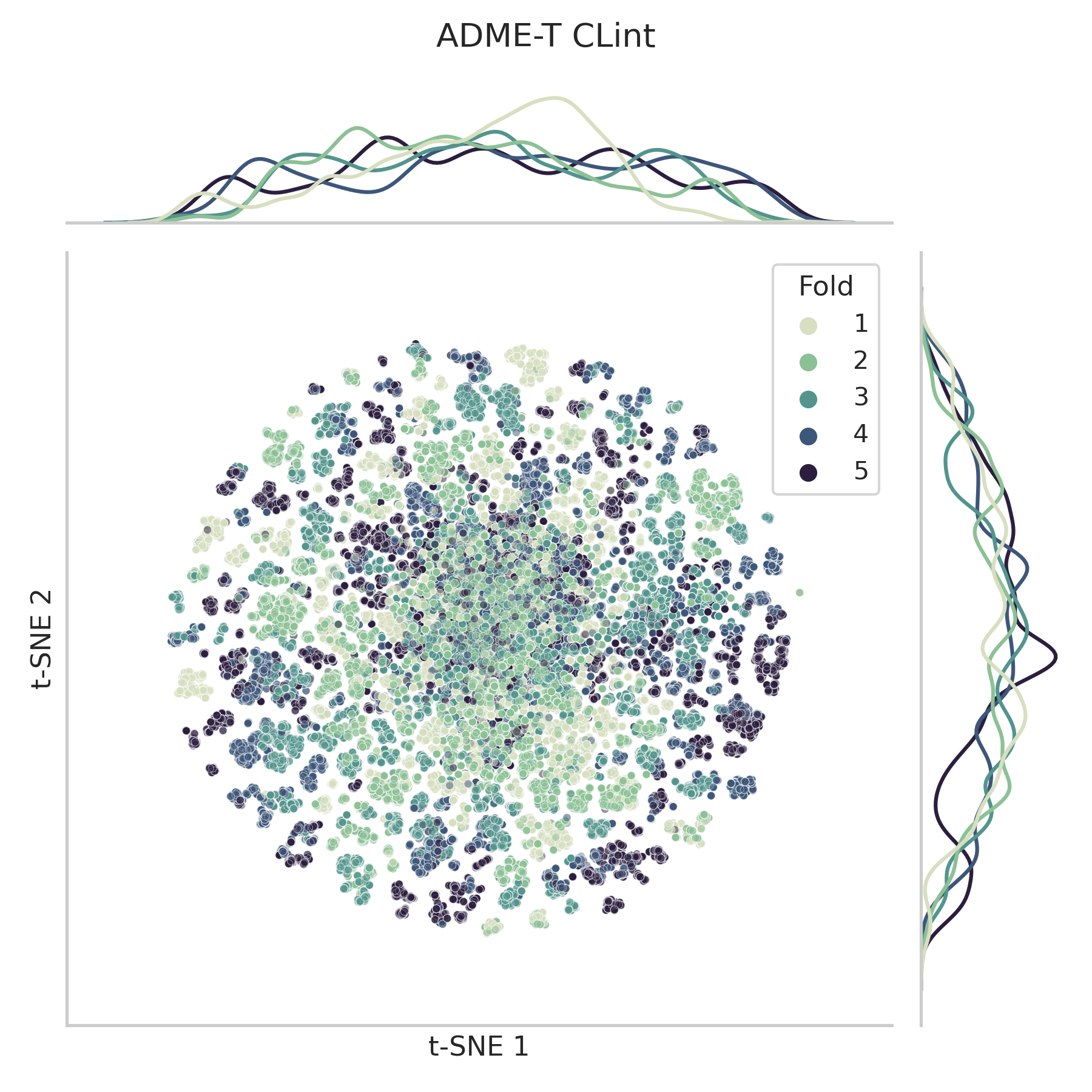}}
        \subfloat{\includegraphics[width=0.2\textwidth]{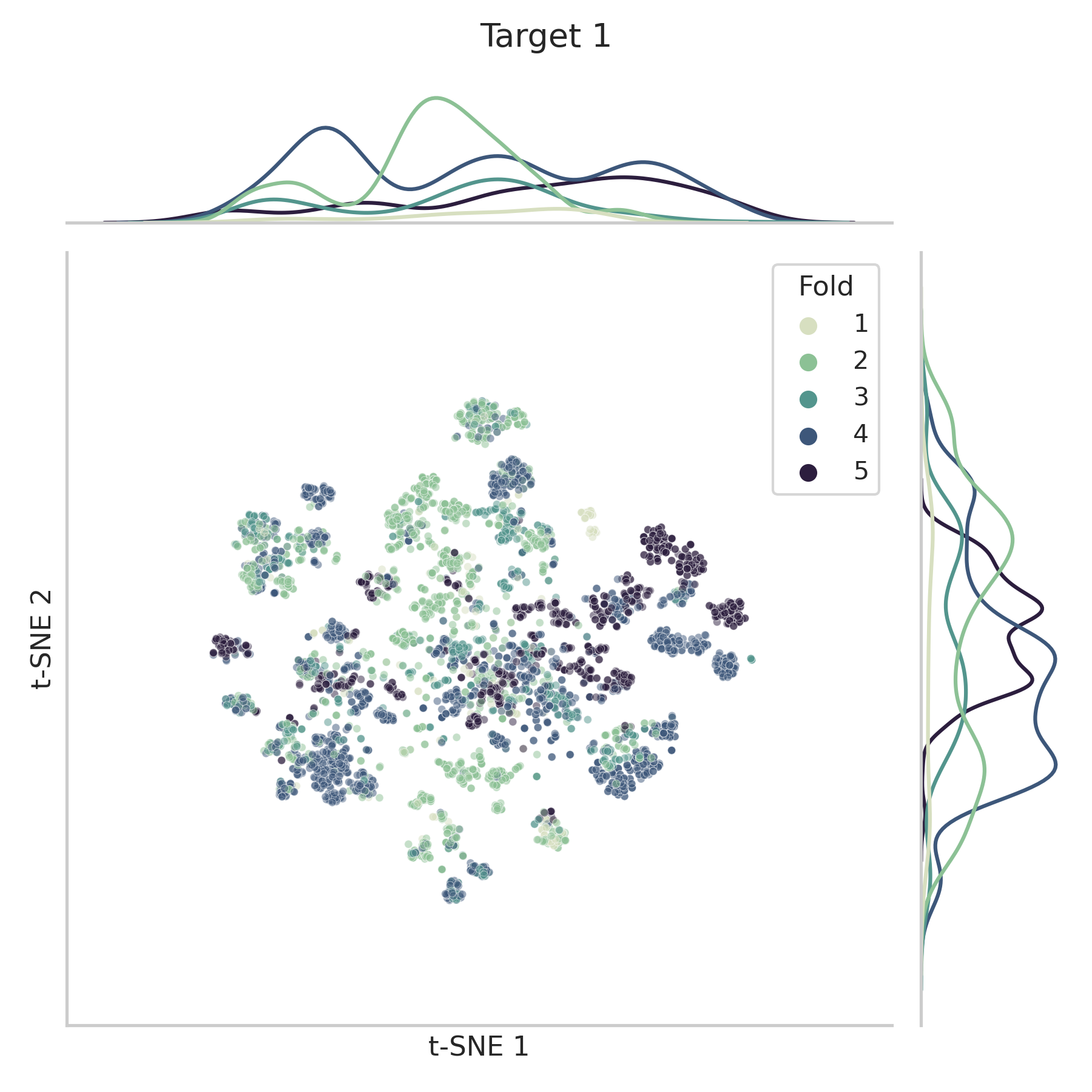}} 
        \subfloat{\includegraphics[width=0.2\textwidth]{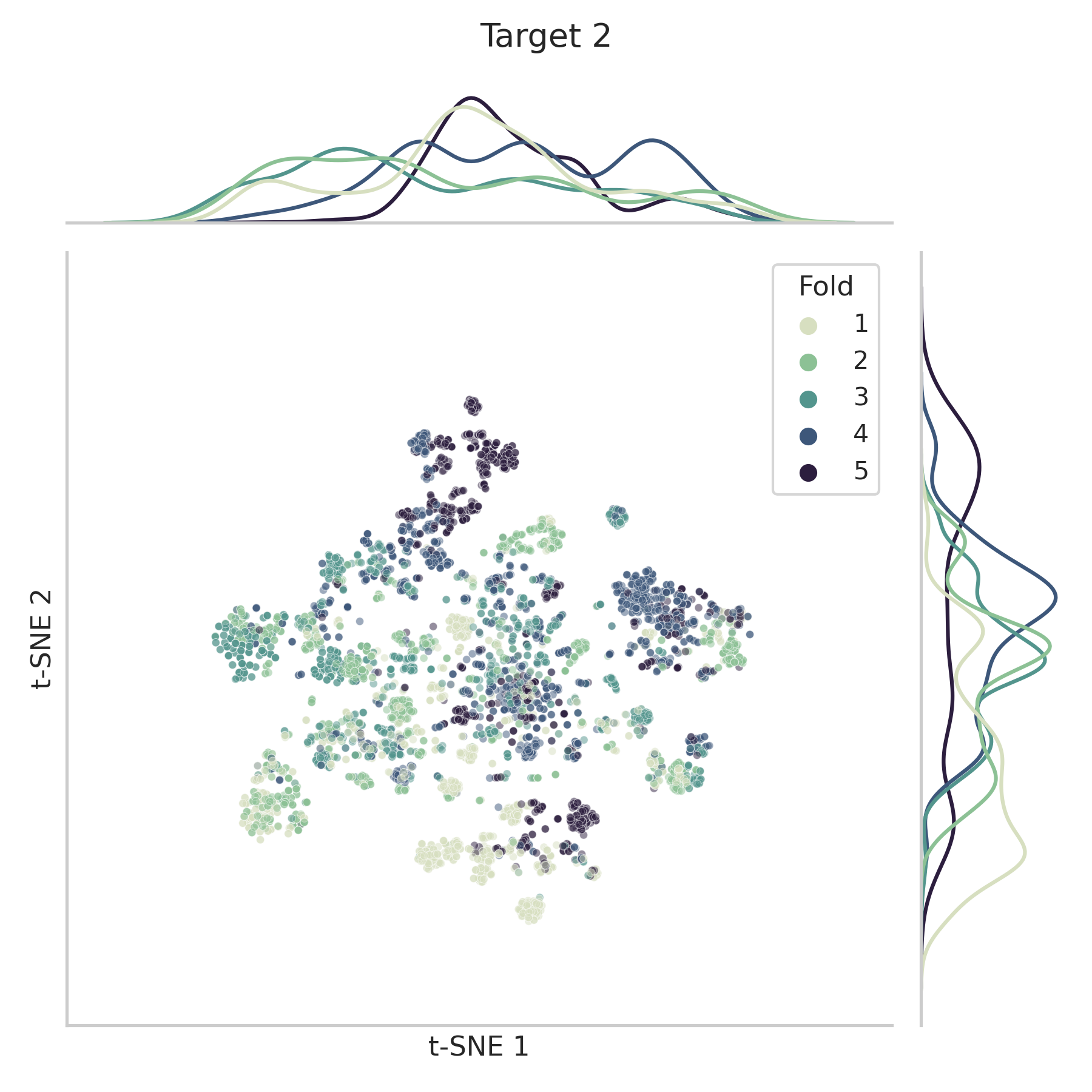}} \\
        \subfloat{\includegraphics[width=0.2\textwidth]{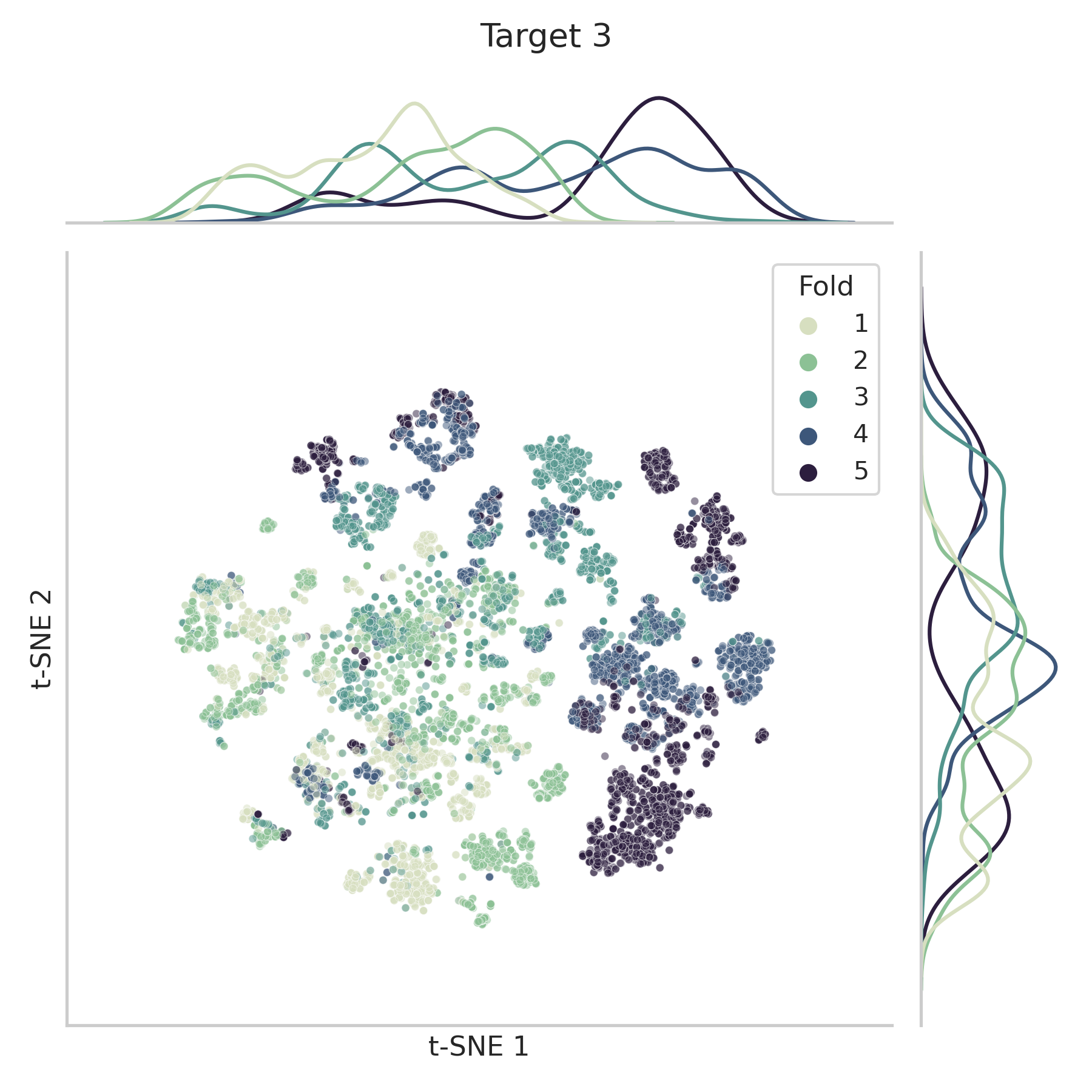}}
        \subfloat{\includegraphics[width=0.2\textwidth]{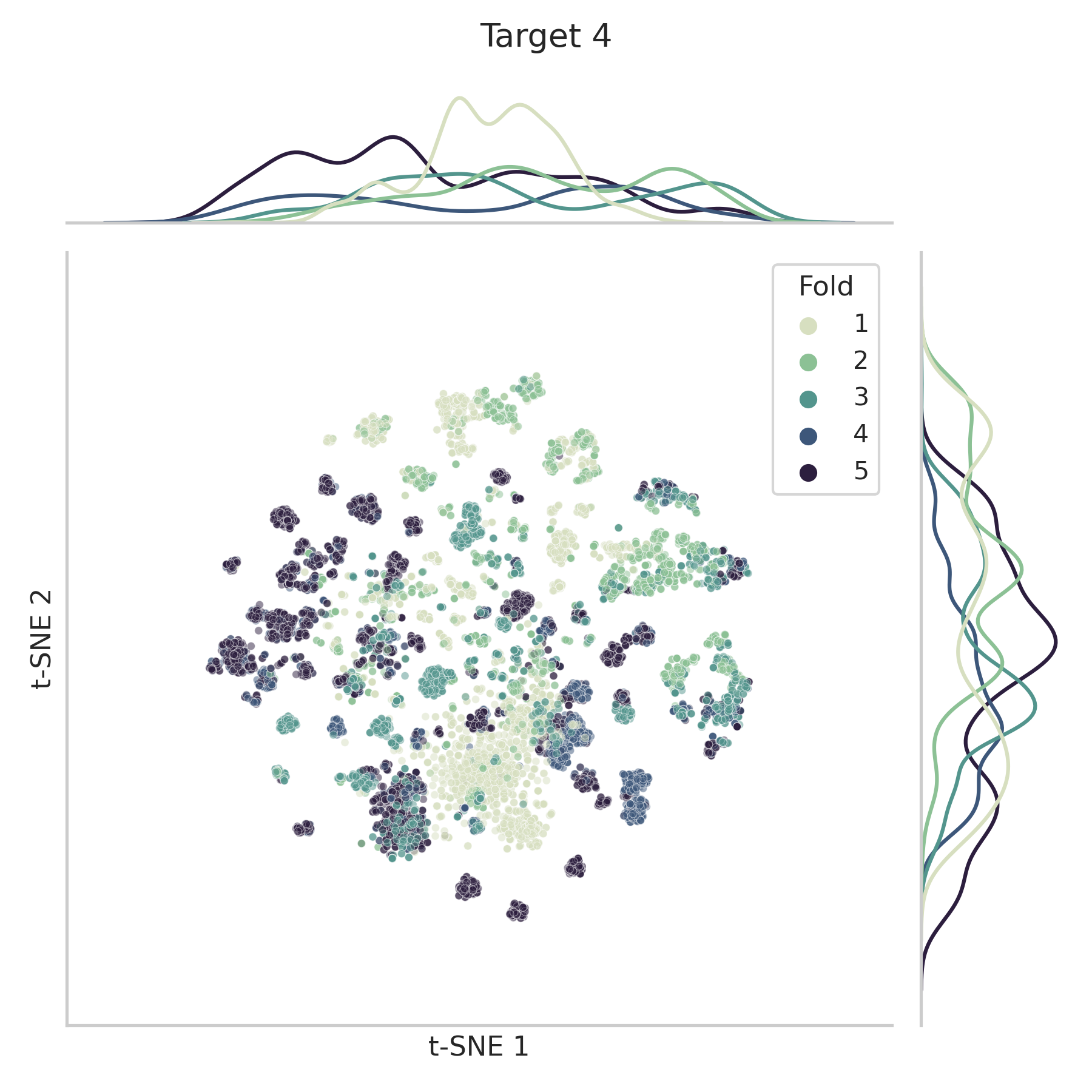}}
        \subfloat{\includegraphics[width=0.2\textwidth]{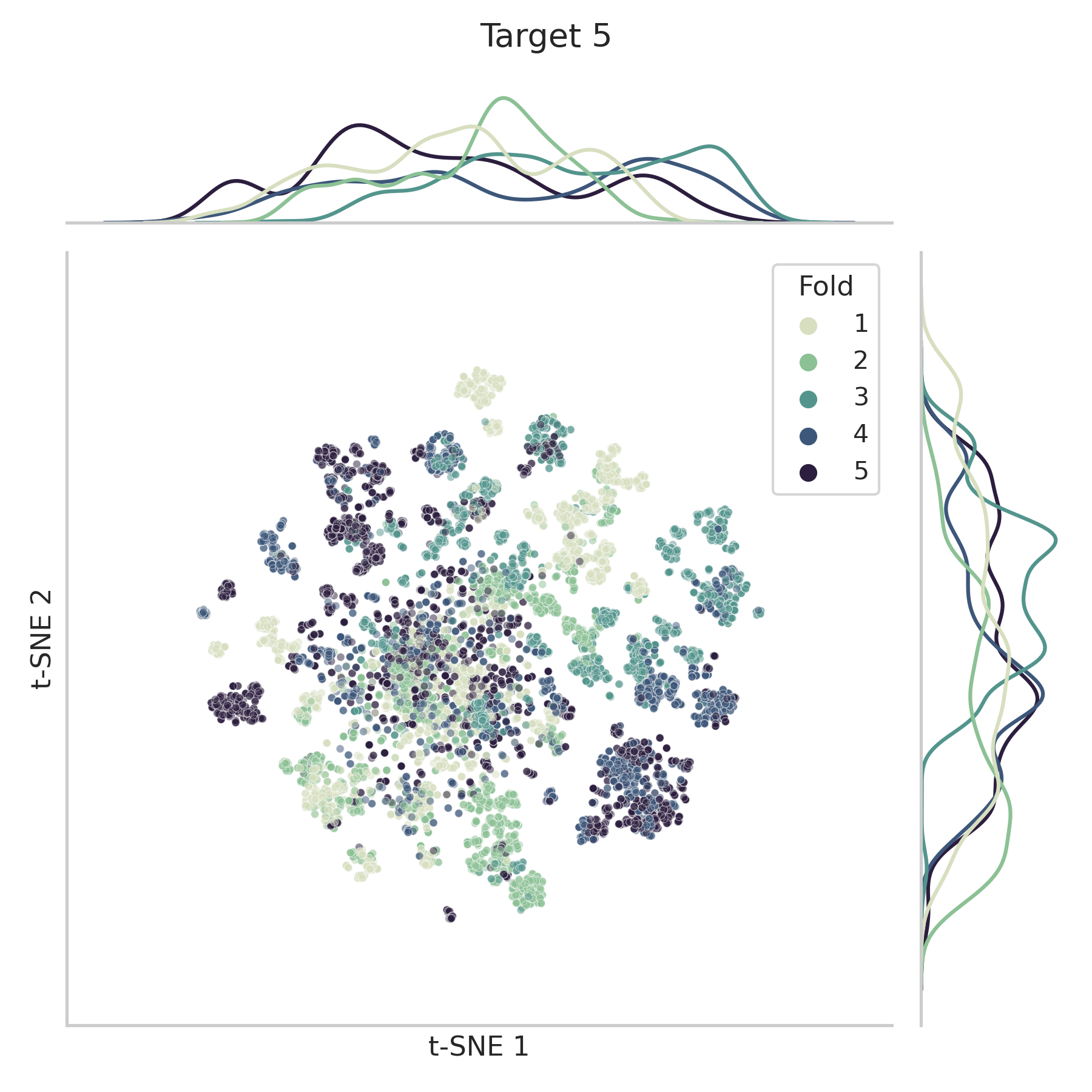}}
        \subfloat{\includegraphics[width=0.2\textwidth]{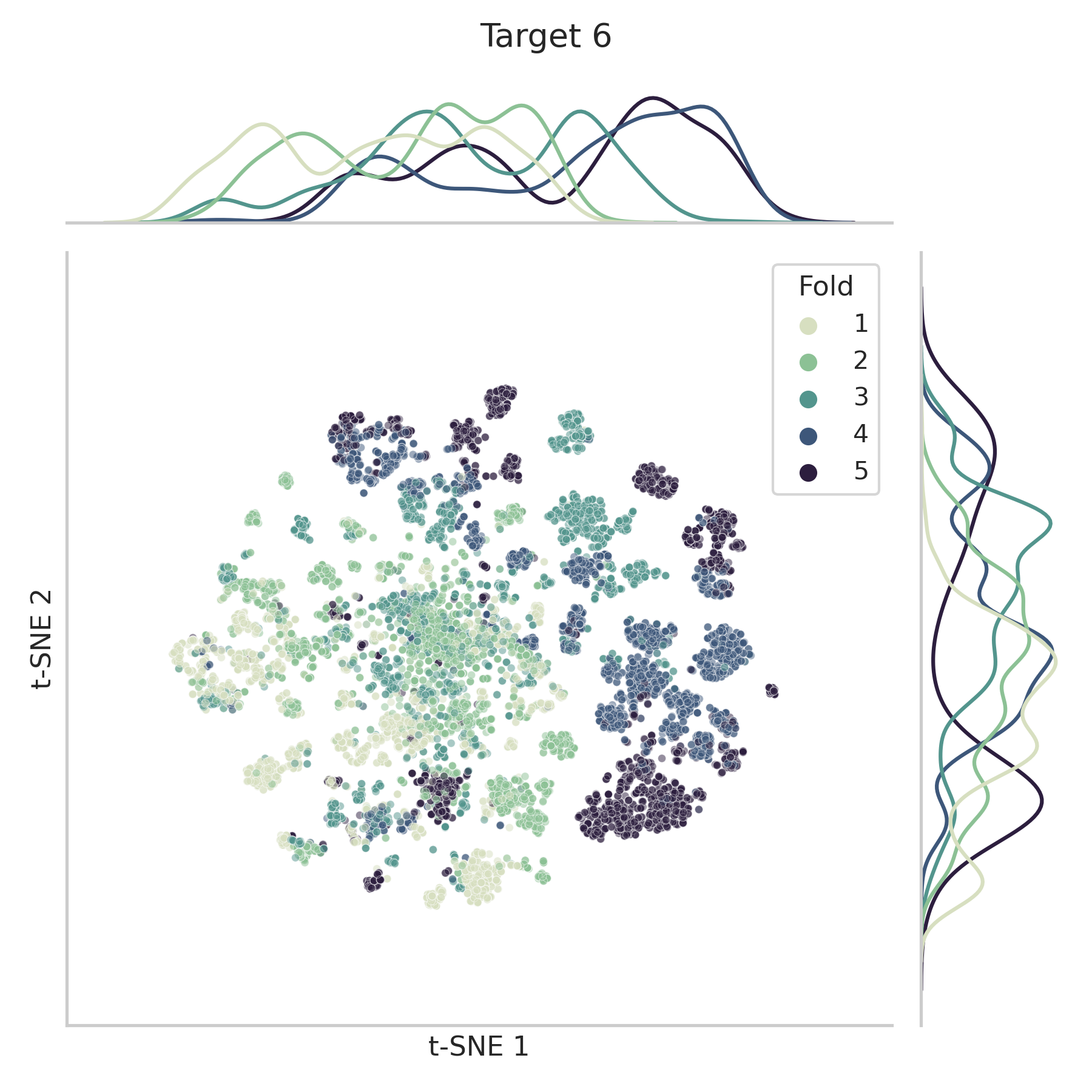}}
        \subfloat{\includegraphics[width=0.2\textwidth]{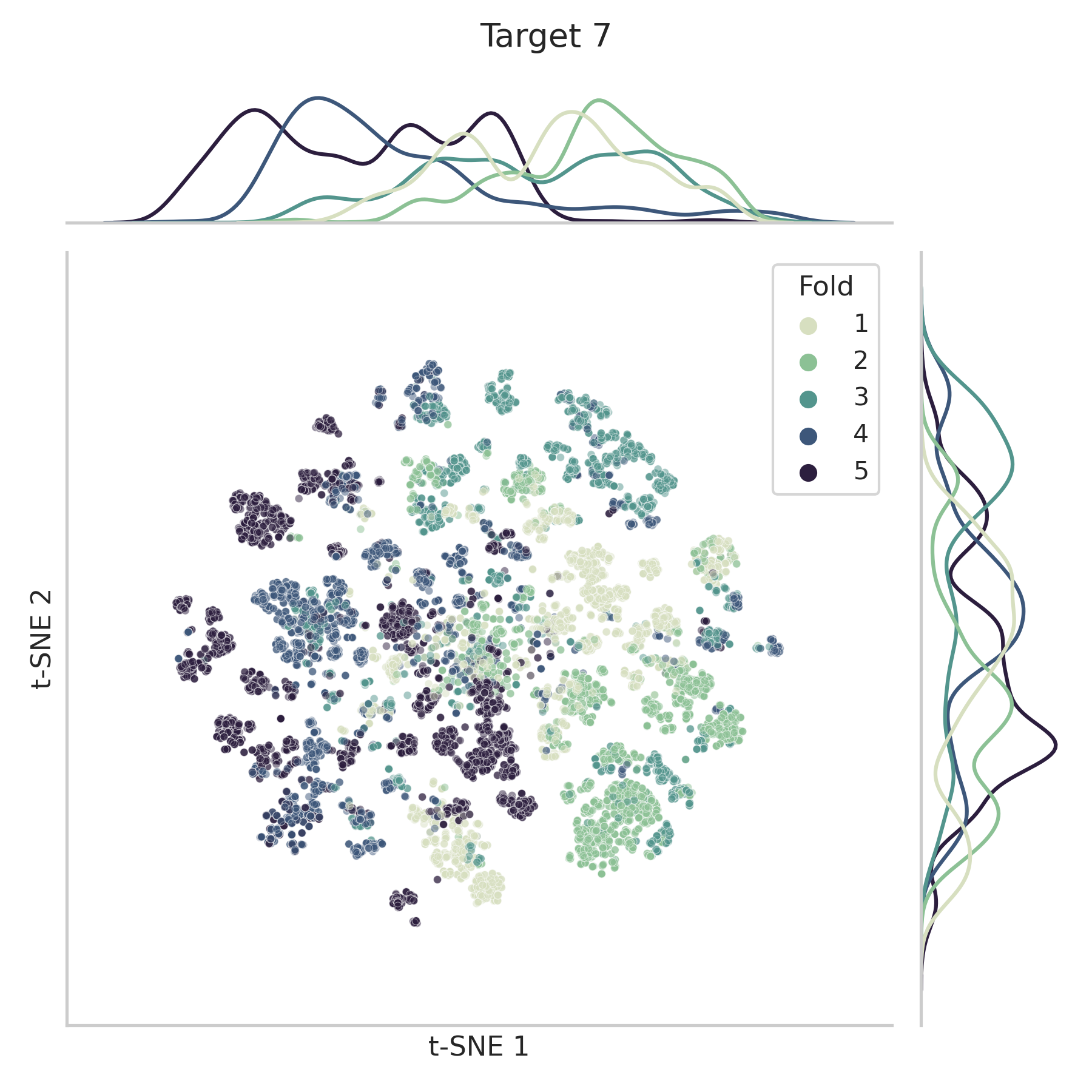}}
        \caption{\textbf{Temporal Distribution of the Feature-space.} Feature space of each assay illustrated in terms of t-SNE projections for each temporal fold.}
        \label{fig:tsne_distributions}
    \end{figure}

    The following section details supporting information about each assay's distribution of the feature space and the label space over time. In Fig.~\ref{fig:label_distributions} the distributions of the observed labels from every assay, are shown for each of the five folds that constitute the three temporal settings illustrated in Fig.~\ref{fig:temporal_split}. Similarly, in Fig.~\ref{fig:tsne_distributions} the feature space is shown as t-SNE projections and colored by the five temporal folds. Generally, ADME-T assays exhibit smaller shifts over time compared to target-based assays. This is true in terms of both the feature space and the label space.  

\section{Model Selection}
\label{app:model_selection}
    For the model selection of all models used in this work, we optimized the hyperparameters described in Table \ref{tab:model_selection} using a grid search according to the loss function of each base model. Each temporal setting was optimized individually. As such, the Random Forest model and the base neural network architecture used in the Ensemble and the MC-Dropout model were all optimized using the validation MSE for observed datasets or the validation CensoredMSE described in Section \ref{sec:methods} for censored datasets. For simplicity, we also used the resulting neural network architecture for the Bayes by Backprop model and the evidential deep learning framework. The base neural network architecture for the Gaussian models was optimized using the validation Gaussian NLL. 
    
    During the grid search for the neural network architectures, we used a maximum of 100 epochs per experiment. However, a maximum of 500 epochs was used for the final experiments, inspired by the neural architecture search proposed by \citet{jiang2024uncertainty}. Additionally, the neural networks were trained using the Adam optimizer with a weight decay of 0.0005, the learning rate was reduced when plateauing with a patience of 50 epochs. The hyperparameter named \textit{decreasing dimension} for the neural networks determines if the hidden dimension is decreased by a factor of two for every layer. Thus, for hyperparameters: 4 number of layers, hidden dimension of size 512, and decreasing dimension True, the resulting hidden layers have sizes [512, 256, 128, 64].

    \begin{table}[ht]
        \caption{\textbf{Model Selection.} Considered hyperparameter space for model selection of Random Forest and the base neural network architectures during grid search based on validation loss.}
            \label{tab:model_selection}
            \vskip 0.15in
        \centering
        \begin{adjustbox}{max width=\textwidth}
        \begin{tabular}{@{}llc@{}}
            \toprule
            Base Model & Hyperparameter & Explored space \\
            \midrule
            \multirow{3}{7em}{Random Forest} & n\_estimators & \{50, 100, 250, 500, 1000\}\\
                                    & min\_samples\_leaf & \{2, 10, 0.25, 0.5, 0.75\}\\
                                    & min\_samples\_split & \{1, 25, 50, 100, 250, 500\}\\
            \midrule
            \multirow{6}{7em}{Neural Network} & Learning rate & \{0.00005, 0.0001, 0.0005, 0.001\} \\
                                        & Scheduler Factor & \{0.1, 0.5\} \\
                                       & Number of hidden layers & \{2, 3, 4\} \\
                                        & Hidden dimension & \{64, 128, 256, 512\} \\
                                        & Decreasing dimension & \{False, True\} \\
                                        & Dropout & \{0.25, 0.5, 0.75\} \\
            \bottomrule
        \end{tabular}
        \end{adjustbox}
    \end{table}

\section{Full Ablation Study}
\label{app:ablation}
\begin{figure}[ht]
        \centering
        \subfloat{\includegraphics[width=\textwidth]{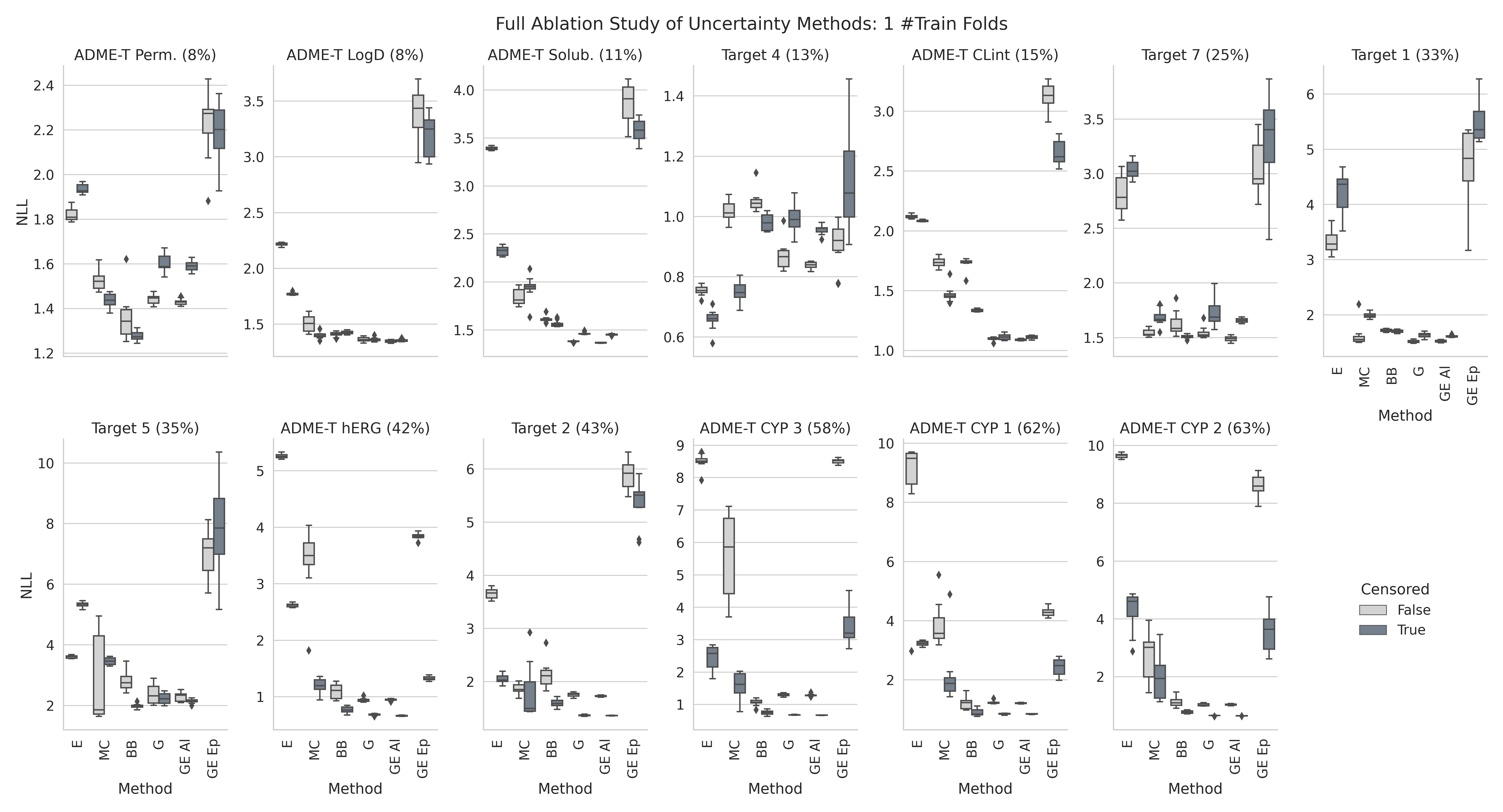}}
        \caption{\textbf{Full Ablation Study: 1 Training Folds.} Comparison of NLL for models trained with and without additional censored labels on the first temporal setting containing one fold for training.}
        \label{fig:full_ablation_1}
    \end{figure}

\begin{figure}[ht]
        \centering
        \subfloat{\includegraphics[width=\textwidth]{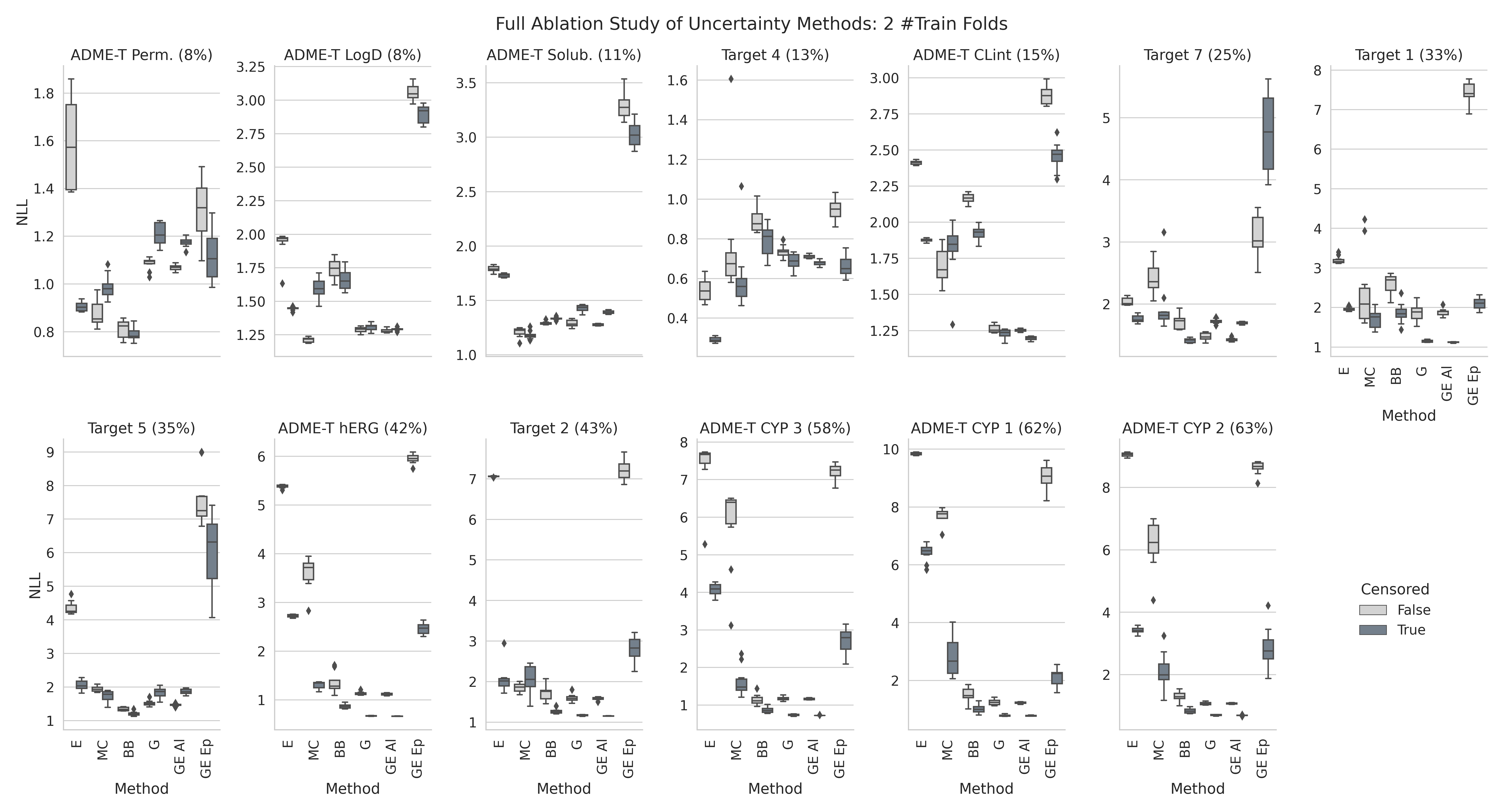}}
        \caption{\textbf{Full Ablation Study: 2 Training Folds.} Comparison of NLL for models trained with and without additional censored labels on the second temporal setting containing two folds for training.}
        \label{fig:full_ablation_2}
    \end{figure}

\begin{figure}[ht]
        \centering
        \subfloat{\includegraphics[width=\textwidth]{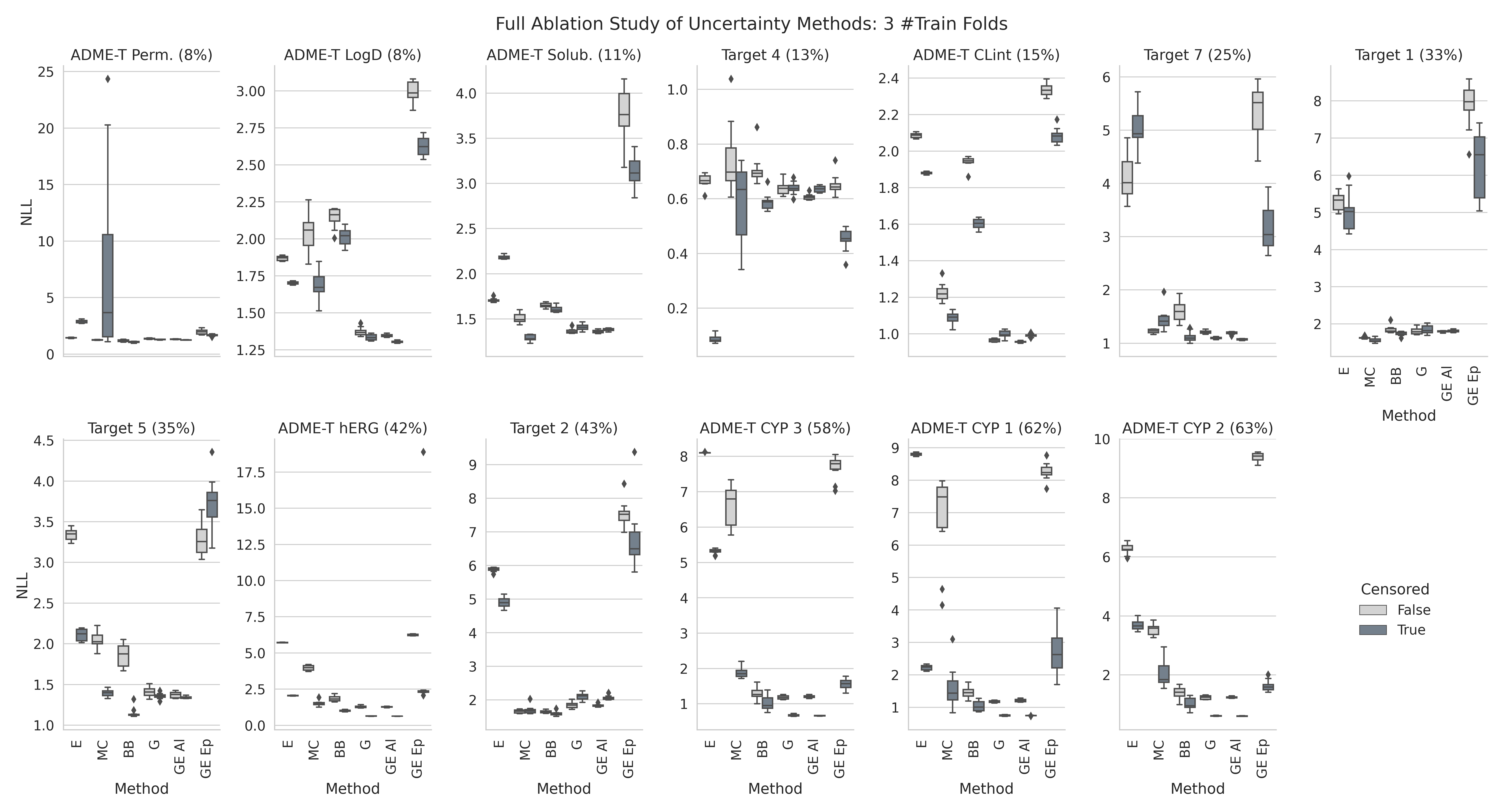}}
        \caption{\textbf{Full Ablation Study: 3 Training Folds.} Comparison of NLL for models trained with and without additional censored labels on the third temporal setting containing three folds for training.}
        \label{fig:full_ablation_3}
    \end{figure}

    Fig.~\ref{fig:full_ablation_1}, \ref{fig:full_ablation_2}, and \ref{fig:full_ablation_3} present the full results from the Ablation study underlying the summary provided in Fig.~\ref{fig:ablation} for each temporal setting respectively. The names of the models have been abbreviated for readability as follows, E for Ensemble, MC for MC-Dropout, BB for Bayes by Backprop,  G for the Gaussian model, GE Al for the aleatoric estimate from the Gaussian Ensemble, and GE Ep for the epistemic estimate from the Gaussian Ensemble. The colors separate the NLL performance for the models trained with and without the additional censored labels. 

\section{Additional Model Comparison}
\label{app:model_comparison}
\begin{figure}[t]
            \centering
            \subfloat{\includegraphics[width=\textwidth]{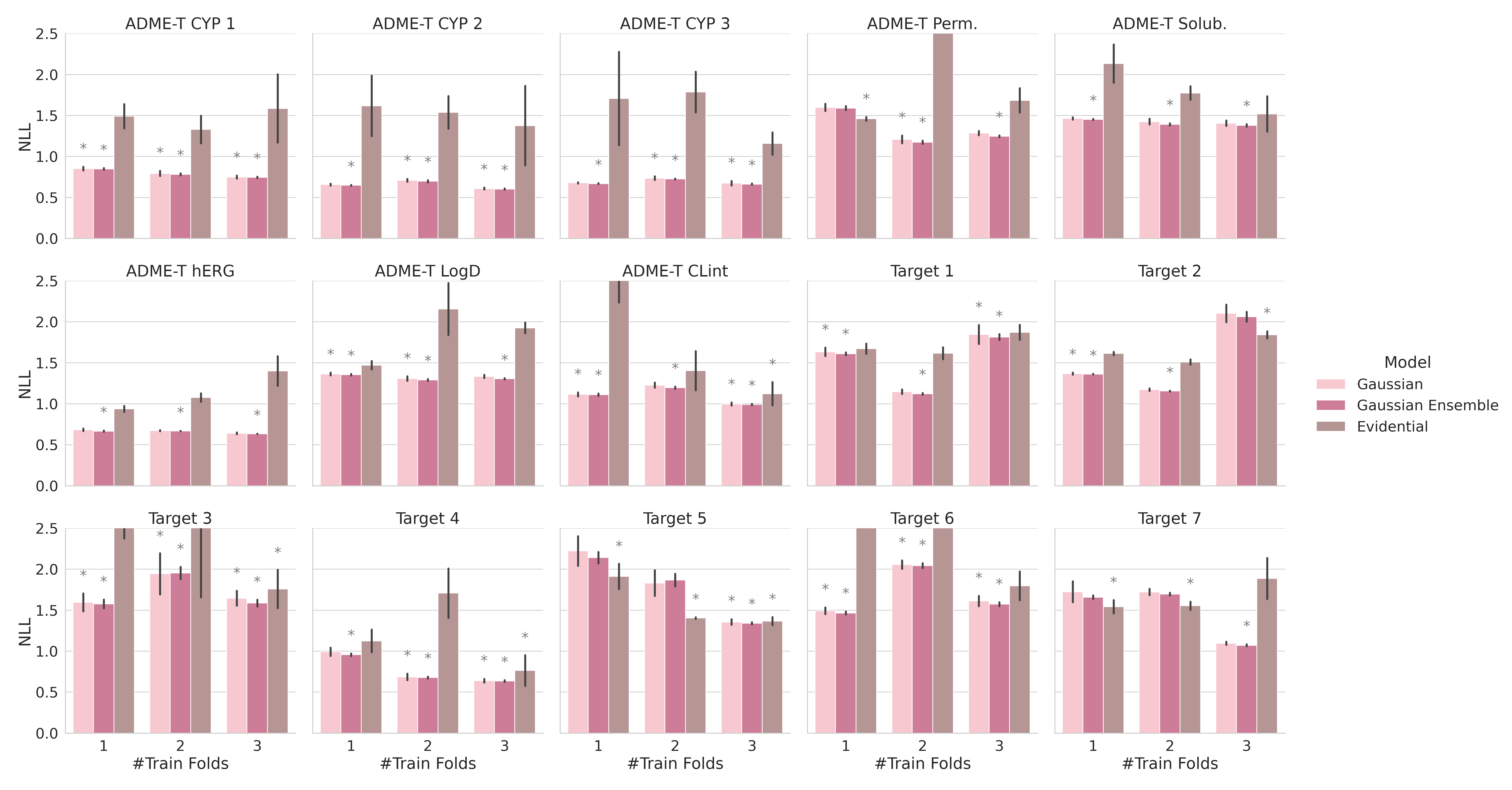}}
            \caption{\textbf{Combined Accuracy of Uncertainty Estimation and Predictive Performance.} Comparing the NLL of all aleatoric uncertainty estimating models, aggregated over 10 experiments. For each dataset, the best model in terms of average NLL is marked with a star together with any other models not statistically worse based on a one-sided Mann-Whitney-Wilcoxon test. Apart from the Random Forest and Evidential models, all other models are trained with censored labels.}
            \label{fig:nll_aleatoric}
        \end{figure}

\begin{figure}[ht]
        \centering
        \subfloat{\includegraphics[width=\textwidth]{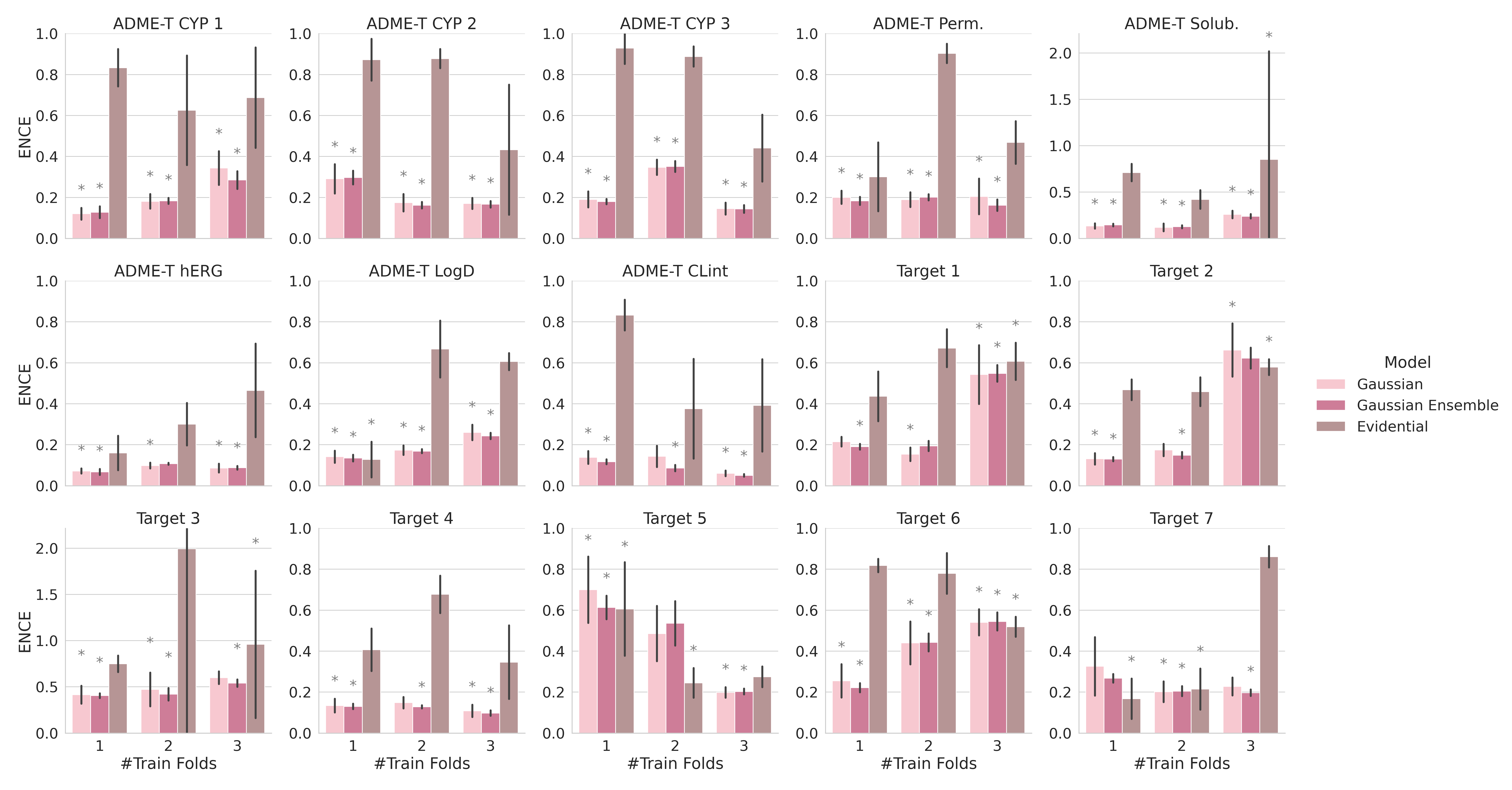}} \\
        \subfloat{\includegraphics[width=\textwidth]{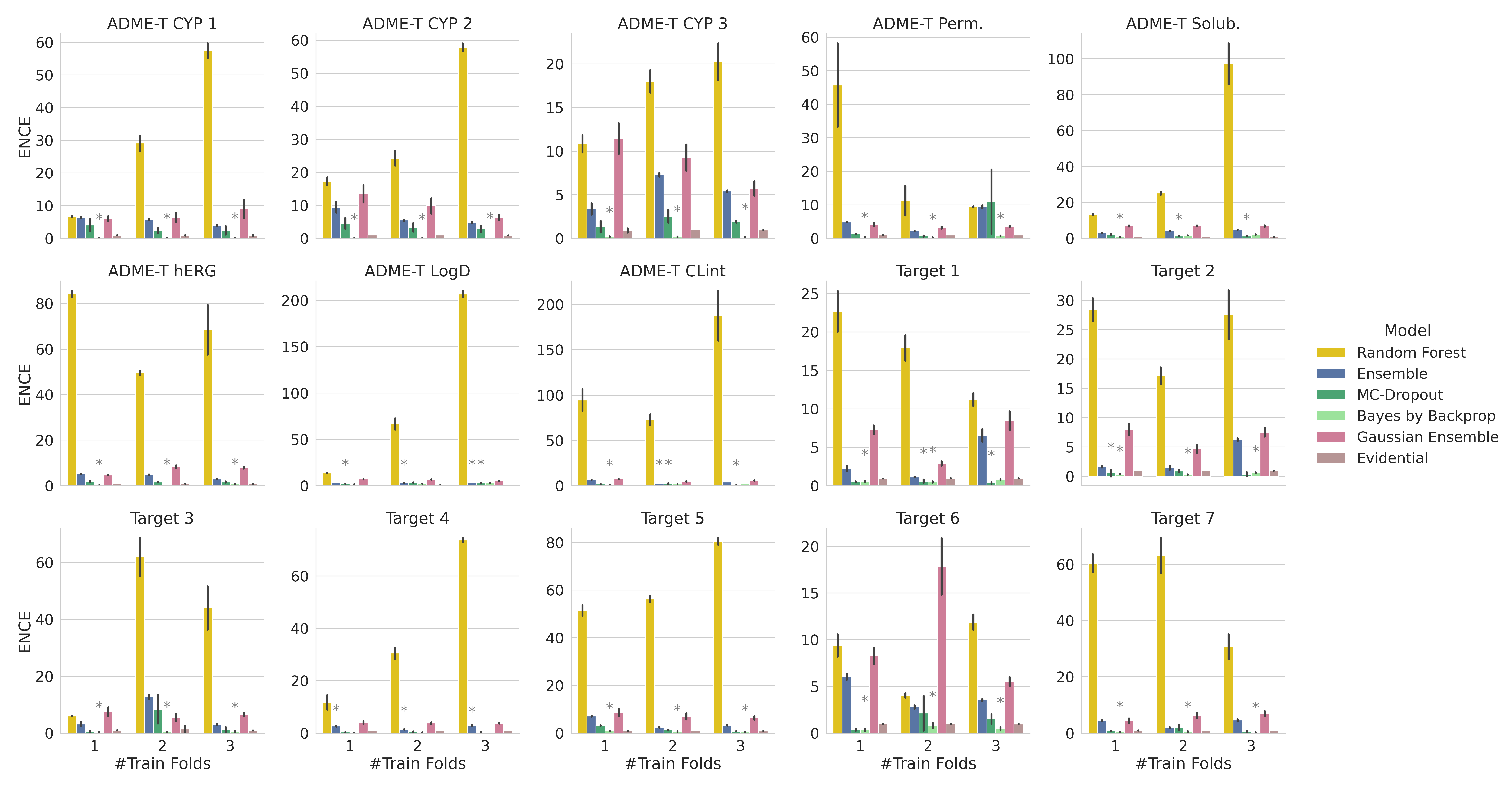}}
        \caption{\textbf{Combined Accuracy of Uncertainty Estimation and Predictive Performance.} Comparing the ENCE of all aleatoric uncertainty estimating models (top) and all epistemic uncertainty estimating models (bottom), aggregated over 10 experiments. For each dataset, the best model in terms of average ENCE is marked with a star together with any other models that are not statistically worse based on a one-sided Mann-Whitney-Wilcoxon test. Apart from the Random Forest and Evidential models, all other models are trained with censored labels.}
        \label{fig:ence}
    \end{figure}

\begin{figure}[t]
        \centering
        \subfloat{\includegraphics[width=\textwidth]{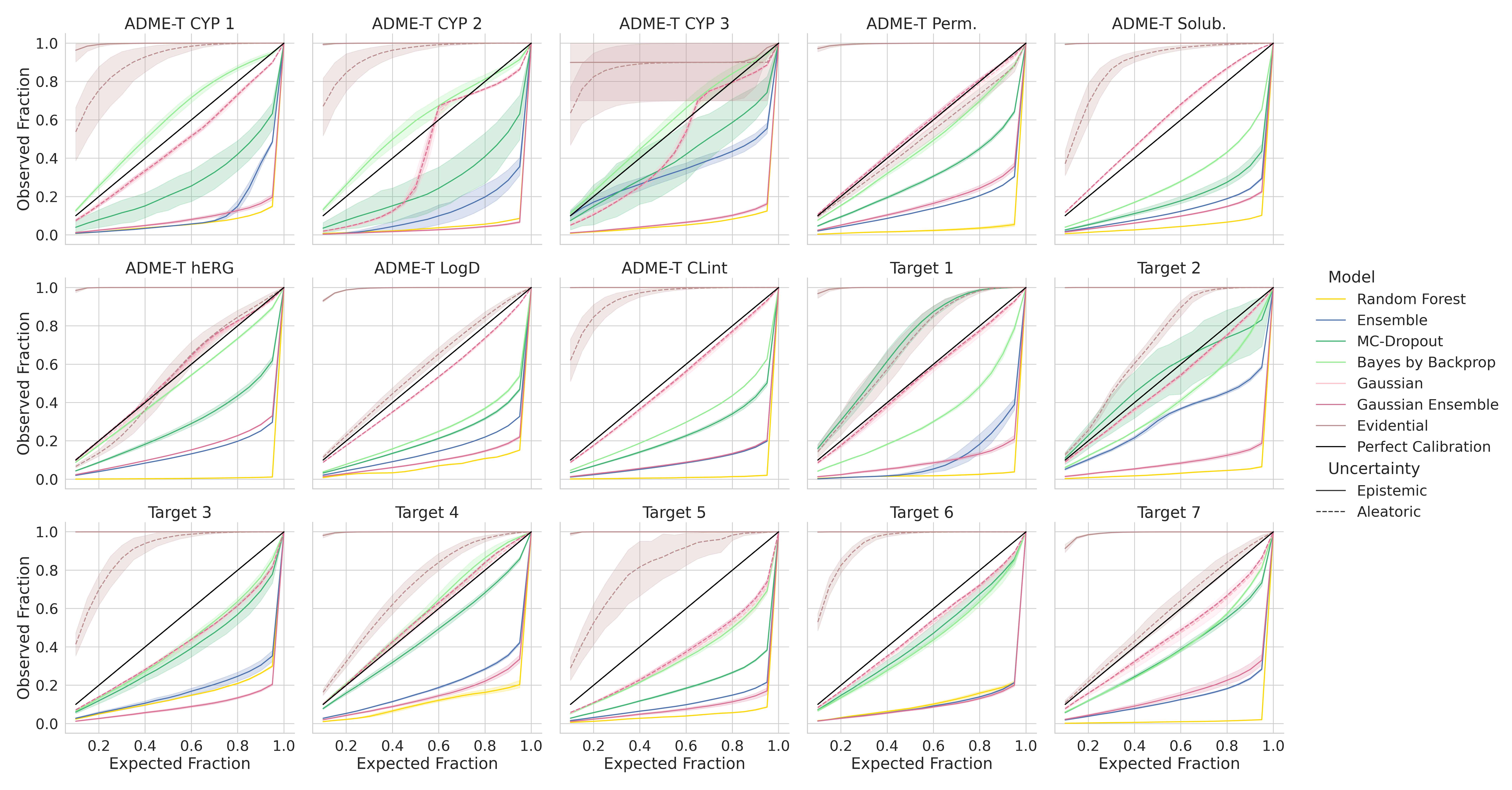}} \\
        \subfloat{\includegraphics[width=\textwidth]{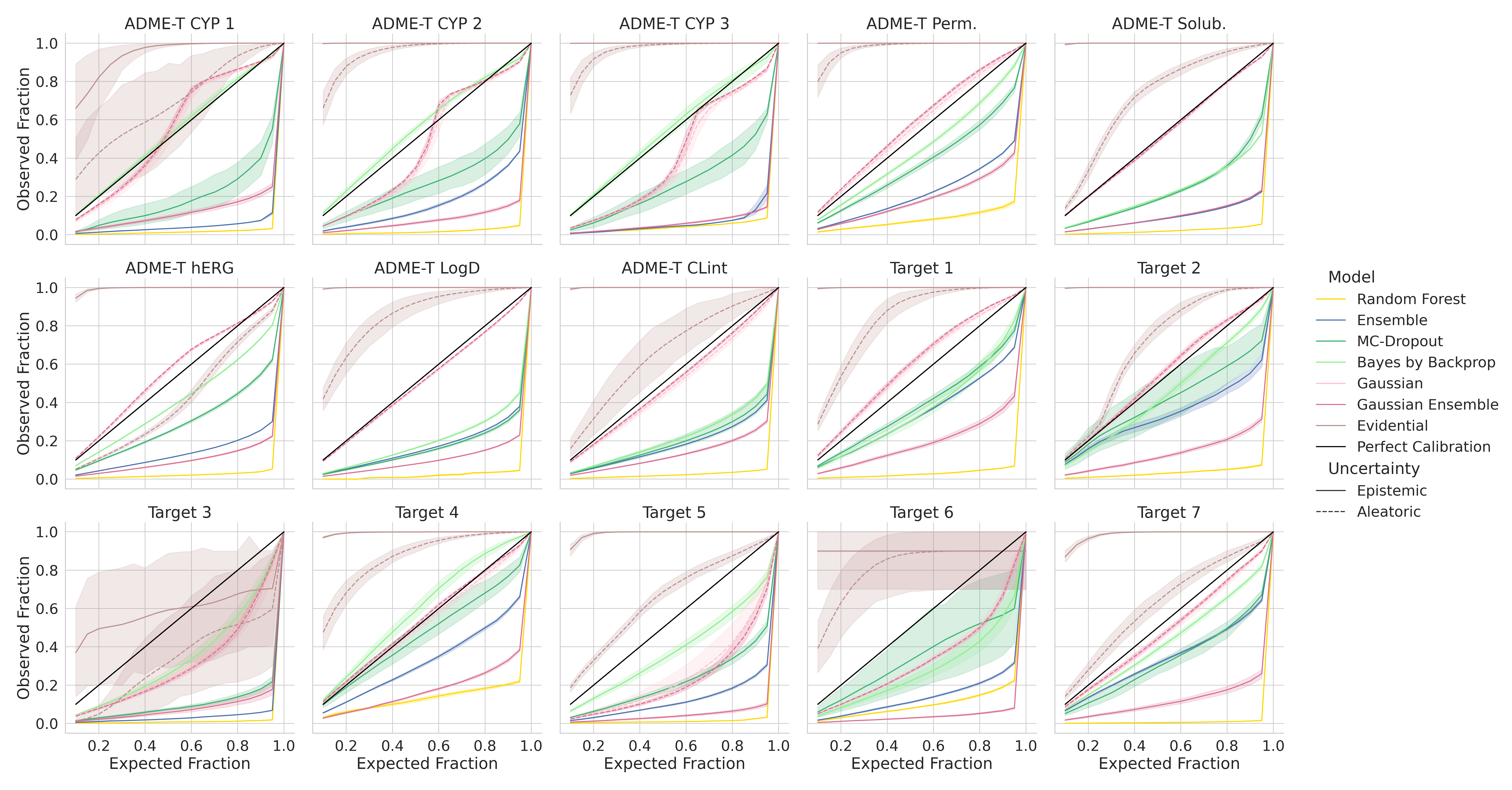}} \\
        \caption{\textbf{Confidence-based Calibration Curves.} Remaining calibration curves for all uncertainty estimates on the first (top) and second (bottom) temporal setting containing one and two folds respectively in the training set, aggregated over 10 experiments. Apart from the Random Forest and Evidential models, all other models are trained with censored labels.}
        \label{fig:calibration_curves12}
    \end{figure}

    In the following section, additional results from the model comparison are presented. First, the NLL scores for all aleatoric estimates are provided in Fig.~\ref{fig:nll_aleatoric}. Next, the ENCE scores are provided in Fig.~\ref{fig:ence} to complement the NLL scores as a second metric for the global, intertwined evaluation of predictive performance and calibration of uncertainty estimates. Next, the confidence-based calibration curves from the first two temporal settings are shown in Fig.~\ref{fig:calibration_curves12}, as a complement to Fig.~\ref{fig:calibration_curves3}. 

\end{document}